\documentclass[fleqn,10pt]{SelfArx} 
\usepackage{preamble_paper}
\usepackage{subfig}
\usepackage{xcolor}
\usepackage{longtable}
\usepackage{xurl}
\usepackage{float}

\newcolumntype{P}[1]{>{\centering\arraybackslash}p{#1}}

\definecolor{color1}{RGB}{0,0,90} 
\definecolor{color2}{RGB}{0,20,20} 

\usepackage{hyperref} 
\hypersetup{hidelinks,colorlinks,breaklinks=true,urlcolor=color2,citecolor=color1,linkcolor=color1,bookmarksopen=false,pdftitle={Title},pdfauthor={Author}}

\JournalInfo{Draft Version} 
\Archive{} 

\PaperTitle{A Novel Method for News Article Event-Based Embedding} 

\Authors{Koren Ishlach\textsuperscript{1}*, Itzhak Ben-David\textsuperscript{2}, Michael Fire\textsuperscript{1}, and Lior Rokach\textsuperscript{1}} 
\affiliation{\textsuperscript{1}\textit{Department of Information Systems Engineering, Ben-Gurion University of the Negev}} 
\affiliation{\textsuperscript{2}\textit{The Ohio State University and National Bureau of Economic Research}} 
\affiliation{*\textbf{Corresponding author}: ishkoren@gmail.com} 

\Keywords{News embedding --- Large scale services --- Deep learning} 
\newcommand{\keywordname}{Keywords} 

\Abstract{
Embedding news articles is a crucial tool for multiple fields, such as media bias detection, identifying fake news, and making news recommendations. However, existing news embedding methods are not optimized to capture the latent context of news events. Most embedding methods rely on full-text information and neglect time-relevant embedding generation. In this paper, we propose a novel lightweight method that optimizes news embedding generation by focusing on entities and themes mentioned in articles and their historical connections to specific events. We suggest a method composed of three stages. First, we process and extract events, entities, and themes from the given news articles. Second, we generate periodic time embeddings for themes and entities by training time-separated GloVe models on current and historical data. Lastly, we concatenate the news embeddings generated by two distinct approaches: Smooth Inverse Frequency (SIF) for article-level vectors and Siamese Neural Networks for embeddings with nuanced event-related information. We leveraged over 850,000 news articles and 1,000,000 events from the GDELT project to test and evaluate our method. We conducted a comparative analysis of different news embedding generation methods for validation. Our experiments demonstrate that our approach can both improve and outperform state-of-the-art methods on shared event detection tasks. 
}

\begin{document}

\flushbottom 

\maketitle 


\thispagestyle{empty} 


\section{Introduction} 
\label{sec_introduction}


In today’s era, many media forms, including online news platforms, television channels, forums, blogs, and podcasts, serve as conduits of information to the public. These media continuously produce and disseminate news, presenting a complex challenge for researchers aiming to understand news impact~\cite{wilding2018impact}. Despite these challenges, multiple studies across various disciplines have employed techniques for analyzing news for diverse purposes. These include analyzing trends in news media \cite{atkinson2014measuring}, detecting bias in news articles \cite{baly2020we}, evaluating the effect of news on the public \cite{ash2020effect}, developing tools to recognize fake news  \cite{MEEL2020112986, naeem2021exploration}, and more. 
For example,  Vasiliki et al.~\cite{9260052} estimated countries’ peace index from the Global Data on
Events, Location, and Tone (GDELT) dataset.\footnote{\url{https://www.gdeltproject.org/}} 
To perform that evaluation, they utilized news documents that were related to events using a machine-coded approach. 
This study is one example of many others who have utilized event-based data to combat real-world issues~\cite{bing2021evolution, consoli2020using, zheng2020comparisons}. 
As Liang Zhao stated "Event analytics are important in domains as different as healthcare, business, cybersphere, politics, and entertainment, influencing almost every corner of our lives"~\cite{zhao2021event}. Moreover, Zhao further claimed that "The analysis of events has thus been attracting huge attention over the past few decades"~\cite{zhao2021event}.


In its simplest form, event data converts natural language reports to a dataset where each entry has the form of \textit{Date, Source\_Actor, Target\_Actor, Event\_Code}~\cite{schrodt2000analyzing}. 
Historically, many researchers conducted their studies using human-coded data, such as the Conflict and Peace Data Bank (COPDAB), which is a longitudinal computer-based library of daily international and domestic events or interactions~\cite{azar1980conflict}, and the WEIS Project dataset, which is a record of the flow of action and response between countries as reflected in public facts~\cite{mcclelland1967world}. 
However, these methods relied on costly human coding efforts that, after decades of research, came to an end~\cite{leetaru2013gdelt}.
As a result, event data started to be generated through content analysis using pattern recognition and simple grammatical parsing methods, making it possible to develop automated coding systems. An example of such a system is Conflict and Mediation Event Observations (CAMEO), a widely used taxonomy~\cite{schrodt2012cameo}.  
The advancement toward automatic coding systems allowed researchers who are interested in employing event data to study international behavior and led to the development of computational methods for the analysis and prediction of various social, political, and economic studies~\cite{Clark_Lax_Rice_2015, blair2020forecasting}. 
However, the automated machine-coded approaches are far from perfect as they are highly restrictive and still suffer from low accuracy, as suggested by Leetaru and Schrodt~\cite{leetaru2013gdelt} - "Small changes in the text or the parsing method could heavily affect the document's tagged events."

To unlock the potential of news articles for these purposes, the development of meaningful news embeddings became a central challenge in the NLP research field, as suggested by Ma et al.~\cite{ma2019news2vec} "In order to facilitate the performance of downstream NLP tasks, it is rather essential to find an efficient
way to represent news as continuous vectors."
Over the years, multiple studies from different domains have utilized news embedding in their work, some notable examples of such works are media bias prediction~\cite{baly2020we}, fake news detection~\cite{mehta2022tackling}, and online news recommendations~\cite{10.1145/3383313.3418477}. 

Two main approaches to news embedding emerged: The first approach involves considering the complete information of articles. 
The prevailing methods in news embedding usually involve capturing semantic context with Deep Neural network architectures. 
For example, Baly et al. ~\cite{baly2020we} used a BERT encoder to generate its news embeddings for its media bias classification task. The second approach relies on other features, such as topic modeling, NERs, and other semantic information, to approach the problem differently.
News2vec~\cite{ma2019news2vec} is an example of such an approach. In this work, news article embeddings were generated by incorporating
relationship information between articles and events, and using additional topological features in a graph comprised of extracted words and features from the article textual data and associated events.

However, a major limitation of news embedding approaches is the ambiguous manner in which the embedding is constructed. News embeddings that provide good performance on several Natural Language Processing (NLP) tasks do not necessarily succeed in the association of articles to events~\cite{setty2018event2vec}.
Moreover, efficiency plays a major part when constructing online news embeddings, since it is considered a streaming task. 
Most of these works \cite{baly2020we,10.1145/3383313.3418477,raza2022news,zhang2021unbert} rely on textual network embedding methods and complex Deep Neural Networks (DNN), which are shown to be expensive both in memory and computation~\cite{xu2021understanding}.

In this study, we address these limitations by focusing on a specific facet and creating a vector space optimized for event-centric embeddings.
To accomplish our goal, we relied on a combination of Named Entity Recognition (NER) techniques and topic modeling. This allows us to make accurate predictions while using fewer resources and providing clear results.
To accomplish this, we introduce in this work a novel method for News Article Event-Based Embedding that receives as input themes, entities, and tagged events of a large set of articles, and the output is event-based embeddings on the given news article information.

\begin{figure*}[t]
    \centering
    \includegraphics[width=1\textwidth]{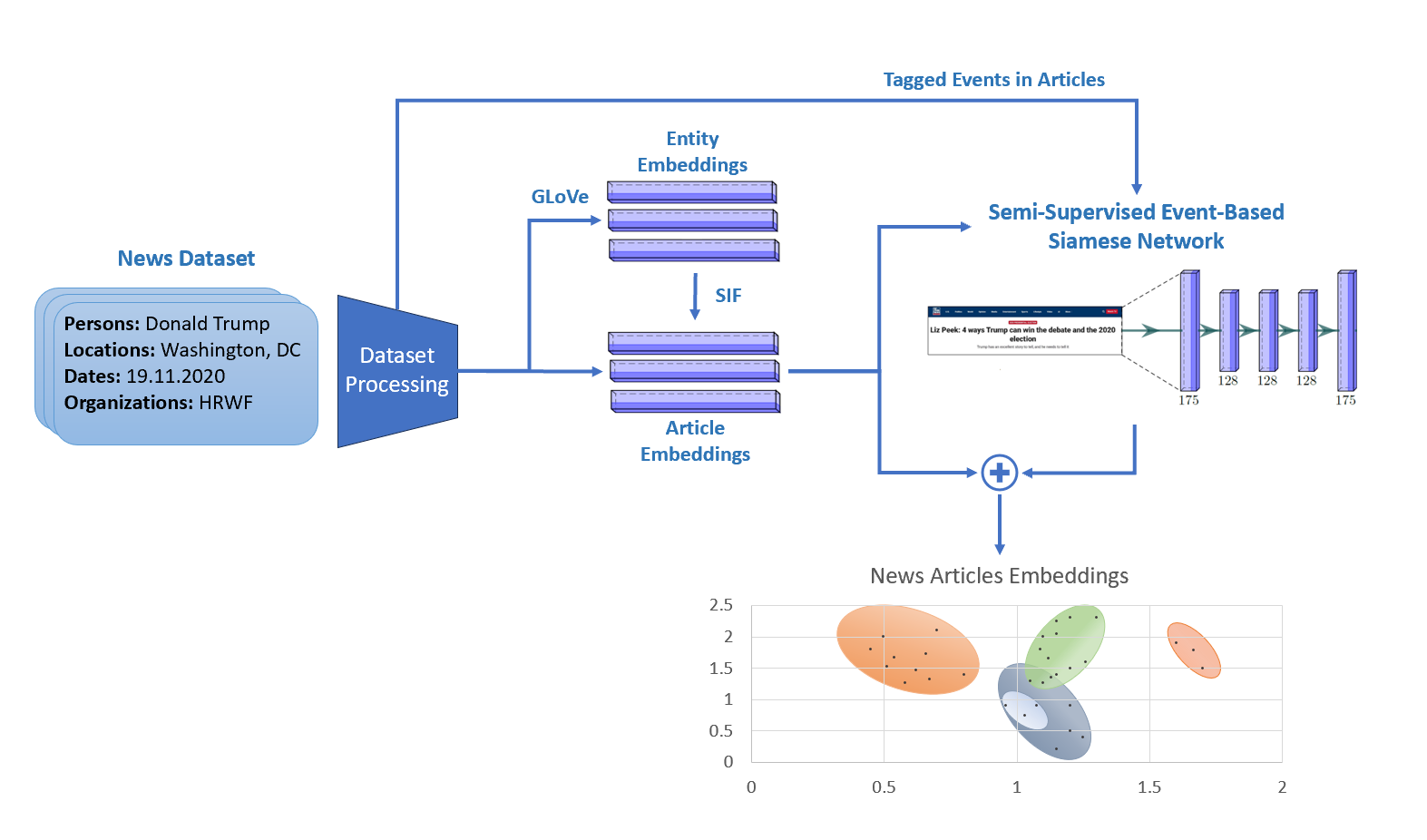}
    \caption[Method Overview]{This figure presents the method's entire pipeline of news embedding generation.}
    \label{fig:pipeline_overview}
\end{figure*}

Our method receives as input the following features of news articles: entities, themes, and tagged events, then it yields as an output event-based news embeddings for the given articles. Our method is constructed in a modular structure of three stages, each stage serves as an input to the next, as can be seen in the method's overview (see Figure~\ref{fig:pipeline_overview}).

\begin{enumerate}
    \item \textit{Articles and Entities Processing}:  A large dataset of articles and entities is processed by executing multiple tasks, including removing redundancies, eliminating article duplication, applying source-based filtering, and entity parsing and removal. 
    \item \textit{Entities Embedding Generation}: Given the processed dataset, we trained separate word embedding models on current and historical articles to create informative embeddings for entities and themes. In this stage, each entity and theme receives a vector representation based on their co-occurrences with each other.
    \item \textit{Articles Embedding Generation}: Given the entity embeddings, we deploy a Smooth Inverse Frequency (SIF) as a methodology to construct document-level vectors (see Section~\ref{sec_related_work}). 
    Then, these embeddings traverse Siamese Neural Networks trained to minimize the distance between articles that share common events.
\end{enumerate}

To test and evaluate our study, we deployed our algorithm on the GDELT project, by collecting five years worth of online news article information from a predefined set of media sources across the United States. After processing, the data contained more than 850,000 unique articles and more than one million events.
We tested our hypothesis by measuring the proposed models' ability to predict whether a pair of articles discuss common events in two different settings: daily test sets and monthly test sets.
We measured both the Precision-Recall Area Under the Curve (PR AUC) metric and the Receiver Operating Characteristic Area Under the Curve (AUC). 
The results demonstrated a significant difference between our suggested semi-supervised method, which relied on event-tagged historical data, and the baseline approach.
On daily aggregated test datasets, our approach yields 0.572 PR-AUC, while the baseline yields 0.546 on average. 
On a monthly aggregated test datasets, our best approach yields 0.369 PR-AUC, while the baseline yields 0.347 on average.

This innovative approach bridges the gap between semantic and event-based context representation, providing a versatile tool for a wide range of applications requiring nuanced event-centric news embeddings.
\\[1\baselineskip]
To summarize, our main contributions are:
\begin{itemize}
    \item We propose a novel way to learn latent features for news articles by extending the Semi-Supervised training process of a Siamese network model to consider article-events relationships.


    \item We provide access to the research community for time-relevant embeddings of more than 320,00 distinct themes and entities. 
    Time-relevant entity embeddings were generated, depending on periodical corpora that purposely overrepresent recently published news articles.
    
    \item We suggest a new approach for evaluating the performance of news embedding methods: pair-wise articles’ event attribution, which utilizes diverse real-world data on relationships between articles and events. 
\end{itemize}

The remainder of this paper is structured as follows. 
The related work in this field is presented in Section \ref{sec_related_work}. 
The proposed method is explained in detail in Section \ref{sec_method}. 
Our datasets, experiments, and final results are presented in Section \ref{sec_experiments}. 
Finally, Section \ref{sec_conclusion} concludes this paper and provides the prospect of future work. 


\section{Related Work}
\label{sec_related_work}
This section contains the background necessary to understand the building blocks and motivation of the research. We discuss previous works in NLP embedding, topic modeling, named entity recognition (NER), and studies on the GDELT project.

\subsection{NLP Embeddings}
Representing words and documents is a crucial part of most, if not all, Natural Language Processing (NLP) tasks. It has been found useful to represent them as vectors, which can be easily interpreted, used in various operations (such as addition, subtraction, distance measures, etc.), and compatible with many machine learning algorithms and strategies.
Our method for constructing news vector representations is influenced by previous research in this field \cite{pennington2014glove, arora2017simple, wang2016learning}.

\subsubsection{Word Embeddings Methods}
Word embedding methods represent words as continuous vectors in a low-dimensional space that captures words' semantic and lexical properties, as stated by Agrawal et al.~\cite{7817051}. 
The representation of words as numerical vectors has been around for more than forty years; however, only recently have neural networks been employed for this purpose.
These embeddings can be obtained from the internal representations of neural network models trained on textual data, as suggested in several works \cite{incitti2023beyond}.

In 2013, Mikolov et al.~\cite{mikolov2013efficient} suggested two iteration-based methods: 
The first is the Continuous-Bag-of-Words (CBOW) model, which predicts the center word from its surrounding context.
The second approach is the skip-gram model, which predicts the surrounding context words given a center word. 
It focuses on maximizing probabilities of context words given a specific center word.

They developed approaches that are considered non-contextual models, in which each word corresponds to a single vector, regardless of how the word is used.
However, recent researchers suggest contextual models in which the training modality allows more vector representations for a word based on how it is used. 
For example, the word "mean" will be represented with a single vector for non-contextual models regardless of its meaning: "average" or "not nice" \cite{incitti2023beyond}.
In 2018, Devlin et al.~\cite{devlin2018bert} work is an example of a contextual model, which introduced the Bidirectional Encoder Representations from Transformers (BERT) model. 
BERT is designed to pre-train deep bidirectional representations from the unlabeled text by joint conditioning in both the left and right contexts in all layers. 

In this work, we utilized the GloVe model developed by Pennington et al.~\cite{pennington2014glove}, relying directly on the low-rank approximation of co-occurrences of themes and entities.
Pennington et al. proposed an efficient method of leveraging statistical information by training only on the nonzero elements in a word-word co-occurrence matrix, instead of the entire sparse matrix or individual context windows in a large corpus. Their results showed that GloVe outperformed Word2Vec and was more efficient, stating: "the efficiency with which the count-based methods capture global statistics can be advantageous" ~\cite{pennington2014glove}.
Since we did not use the full text of news articles, GloVe is well-suited for this work's use case. Instead, we adopted an efficient approach that focuses on themes and entities while ignoring their position and textual context.

\subsubsection{Sentence and Document Embeddings Methods}
Sentence and document embedding methods map a feature vector representing a document or text. 
For instance, previous works have computed phrase or sentence embeddings by composing word embeddings using operations on vectors and matrices~\cite{le2014distributed}. 
Another example of such an approach is proposed by Phang et al.~\cite{phang2018sentence}, as they trained BERT, a powerful pre-trained sentence encoder. 

In this study, we utilized the Smooth Inverse Frequency (SIF) algorithm for a naive representation of news article vector representations \cite{arora2017simple}.  SIF computes sentence embeddings as a weighted average of word vectors by deploying a weighting scheme known as Smooth Inverse Frequency, which is meant to improve sentence embedding performance.  SIF aimed to capture the significance of words in a sentence by down-weighting common words and up-weighting less frequent words~\cite{arora2017simple}.

\subsection{Topic Modeling}
Understanding the core themes associated with a document collection is a fundamental task in today's information era~\cite{churchill2022evolution}.
The purpose of topic models is to detect those "themes," which are the ideas and subjects that "connect" a set of text documents. An example of a possible theme is "police brutality."
These models are a class of unsupervised machine-learning techniques designed for this task. 
In this study, we define a topic model as an unsupervised mathematical model that takes as input a set of documents $D$ and returns a set of topics $T$ that represent the content of $D$ accurately and coherently as suggested by Kherwa and Bansal~\cite{kherwa2019topic}.

Churchill and Singh's survey \cite{10.1145/3507900} cited several studies that used topic models to analyze climate change literature \cite{sleeman2017modeling} and to understand the 2016 US presidential election through newspapers \cite{bode2019words}. Additional works focused on developing new methods for topic modeling and evaluating their performance on news-based datasets, such as the 20 Newsgroups dataset \cite{nguyen2015improving, moody2016mixing}.\footnote{Link for the 20 Newsgroup dataset: \url{http://qwone.com/~jason/20Newsgroups/}}

In our work, we decided to use the Global Content Analysis Measures (GCAM) system, which runs each news article monitored by GDELT through an array of content analysis tools to capture over 2,230 latent dimensions, reporting density and value scores for each. 
In the overview of the GDELT project, we provide further information on their approach \ref{gdelt_subsection}.

\subsection{Named Entity Recognition}
Named Entity Recognition (NER) is a versatile tool that enables the extraction of information from various sources. 
It is an essential component of Natural Language Processing (NLP) applications such as text comprehension \cite{zhang2019ernie}, information retrieval \cite{guo2009named}, machine translation, and knowledge base construction \cite{kejriwal2019domain}.

The most commonly used typologies consist of a few high-level classes such as person, organization, and location \cite{ehrmann2016named}. 
There is a variety of open-source code designed for NER tasks, such as spaCy \cite{vasiliev2020natural}, NLTK \cite{hardeniya2016natural}, and flair \cite{akbik2019flair}. 
A NER system aims to learn patterns from labeled examples, which can be used to classify new sequences of tokens.

There are mainly three main approaches for performing NER: rule-based, machine learning, and deep learning.
Rule-based systems are interpretable and do not require training data, however designing the rules can be time-consuming and requires expertise.
For example, Diez et al.~\cite{diez2021medieval}, developed a rule-based system to recognize Medieval Spanish person names from various genres of manually transcribed texts, by defining a custom entity typology of eight main types and used multiple modules to recognize names and person attributes.

Machine-learning NER systems are statistical models that are built upon machine-learning algorithms through annotated data and selected features.
Ehrmann et al. claimed in their survey~\cite{10.1145/3604931} that "Heavily researched and applied in the 2000s, machine learning-based approaches contributed strong baselines for mainstream NER, and were rapidly adopted for NER."

Ehrmann et al. also claimed that deep learning techniques are dominating NER developments~\cite{10.1145/3604931}.
These deep learning approaches mainly rely on sequence labeling and use BiLSTM architectures or self-attention networks to learn sentence or sequence features~\cite{10.1145/3604931}.

Previous studies have tried to stack multiple LSTMs for sequence-labeling NER as they follow the trend of stacking forward and
backward LSTMs independently, the Baseline-BiLSTMCNN is only able to learn higher-level representations of past or future.
Chiu et al.~\cite{chiu2016named} presented an architecture that detects word and character-level features using a hybrid bidirectional LSTM and CNN. 
It removes the need for most feature engineering and proposes a novel method of encoding partial lexicon matches in neural networks. 

In our research, we relied on the NER algorithm output deployed by the GDELT project as input.
They deployed three engines for recognizing emotions, dates, persons, organizations, and locations by identifying an array of other kinds of proper names while attempting to compile a list of all amounts of each entity expressed in each article to offer numeric context to global events.
Further description of this process can be found in the GDELT project codebook.\footnote{\url{http://data.gdeltproject.org/documentation/GDELT-Global_Knowledge_Graph_Codebook-V2.1.pdf}}

\subsection{NLP News Embeddings}
Over the years, researchers developed many supervised and unsupervised methods to represent entities, articles, or news events. 
In 2017, Conneau et al.~\cite{conneau2017supervised} investigated whether supervised learning can be leveraged for sentence embedding representations. 
They investigated the impact of the sentence encoding architecture on representational transferability, and compared convolutional, recurrent, and even simpler word composition schemes.
To evaluate their research, Conneau et al. used news articles embedded in sentences and predicted a similarity score between 0 and 5.

In 2018, Setty et al. presented Event2vec \cite{setty2018event2vec}. They used proposed network embedding techniques for the embedding representation of news events.
As events involve different classes of nodes, such as named entities, temporal information, etc. General-purpose network embeddings are agnostic to event semantics. 
To address this problem, they proposed biased random walks tailored to capture news events' neighborhoods in event networks.

In 2019, Ma et al. presented News2vec \cite{ma2019news2vec}, which represents the context of articles, events, and additional features in a graph.
They focus on both the contextual relationship between news and potential connections as the network becomes more profound, with the embedded vectors containing semantic and labeled information as well as latent connections between different news events.

Since then, most of the works have utilized news embeddings as part of a technique to achieve specific tasks~\cite{mehta2022tackling, 10.1145/3383313.3418477, zhang2021unbert}, rather than suggesting a generic approach for news article embedding.
For example, in 2022, Mehta et al. \cite{mehta2022tackling} work on fake detection included the following methodology: They embedded using graph neural network articles; then they formulated inference operators that augment the graph edges by revealing unobserved interactions between its elements, such as the similarity between documents’ contents and users’ engagement patterns.

However, most of these work for news textual network embedding methods still rely on full-text and complex DNN, which are expensive both in memory and computation, as Xiang and Wang claimed in their event extraction study: ”How to design an efficient neural network architecture is the main challenging issue for event extraction based on deep learning” ~\cite{xiang2019survey}. 
Furthermore, the works that handle the previous limitations usually fail to collect features that do not solely describe the contextual information at the document level, as stated in \cite{ma2019news2vec}: "The major limitation of embedding approaches discussed above is that they produce representations and features that solely describe the contextual information at the document level". 
Another limitation is that in previous works in the field, the evaluation process depends on measuring several tasks that do not necessarily determine the success of associating articles with events. 
In the works of Ma et al. and Setty et al. \cite{ma2019news2vec, setty2018event2vec} that concentrated on the embedding process of news, they chose for evaluation tasks of news recommendation, or stock prediction tasks, as well as the majority of works in the field that utilized news embeddings for a specific task, such as described in the following services \cite{MEEL2020112986, zhao2021event, phan2023fake, raza2022news}.

There are numerous challenges to performing the task of associating events with articles. First, the positive rate is very low. In our case, on average, the probability of two articles sharing a common event given they were published in the same month is less than 0.05\% (see Table \ref{tab:datasets_statistics}).
Second, an article can be associated with multiple events, creating a state of uncertainty in the detection process and limiting the possible solutions. To alleviate this difficulty, some previous works even chose to assume that a single article could only be related to one event \cite{upadhyay2016making}.
Third, as we discussed earlier, robust methods that rely on full textual data, transformers, or graph neural networks often require high computational and memory resources \cite{xu2021understanding, treviso2023efficient}.

\subsection{GDELT Project}
\label{gdelt_subsection}
The GDELT Project \cite{leetaru2013gdelt} is a real-time network diagram and database of global human society for open research which monitors the world’s broadcast, print, and web news from nearly every corner of every country in over 100 languages and identifies the people, locations, organizations, counts, themes, sources, emotions, counts, quotes, and events driving our global society every second of every day, creating a free open platform for computing on the entire world.

By employing named entity and geocoding algorithms \cite{leetaru2012data}, The GDELT Global Knowledge Graph (GKG) compiles a list of every person, organization, company, and location from every news report, explicitly designed for the noisy and ungrammatical world that is the world's news media.
In the GDELT project, tones refer to the emotional sentiment or attitude expressed in news articles and reports. 
The tones are quantified on a numerical scale ranging from extremely negative to extremely positive.
The tones in GDELT data are represented as numerical values, typically ranging from -100 (extremely negative) to +100 (extremely positive), with 0 being neutral.\footnote{\url{https://analysis.gdeltproject.org/module-gkg-tonetimeline.html}}
Moreover, GDELT provides the Global Content Analysis Measures (GCAM),\footnote{\url{https://blog.gdeltproject.org/introducing-the-global-content-analysis-measures-gcam/}} this system runs on each news article monitored by GDELT through an array of leading content analysis tools, and topic models to capture and quantify latent emotional and thematic signals subconsciously encoded in the world’s media.

The GDELT Project offers a comprehensive event database that contains records of more than 300 types of physical activities taking place around the world. 
These activities can range from riots and protests to peace appeals and diplomatic exchanges. The database provides the location of the activity, whether it’s a city or a mountaintop, and covers the entire planet, dating back to January 1, 1979. 
It is updated every 15 minutes, making it a reliable source of real-time information. Examples of some identified events include ”ENVIRONMENT AND NATURAL RESOURCES”, ”LABOR MARKETS” and ”CRISIS WELLBEING HEALTH”.\footnote{\url{https://blog.gdeltproject.org/gdelt-2-0-our-global-world-in-realtime/}}.
GDELT also considers "mentions," which are the unique pairings of an article and the event it discusses. Multiple articles can discuss the same event, and a single article can cover multiple events.

Using GCAM, one can assess the density of “Anxiety” speech via Linguistic Inquiry and Word Count (LIWC)\footnote{\url{https://www.liwc.app/}} project, "Positivity" via Lexicoder,\footnote{\url{https://quanteda.io/reference/data_dictionary_LSD2015.html}} “Smugness” via WordNet,\footnote{\url{https://wordnet.princeton.edu/}} and so forth, in total, 18 content analysis systems totaling more than are deployed. 
The volume of the topic models deployed by GDELT, and the accessibility they provide the output per article, make it an extremely valuable resource for our analysis.


\section{Methods}
\label{sec_method}
The main goal of this study is to develop a method for generating event-centric news embeddings.
To address this goal, we present our lightweight generic method that consists of three stages, as depicted in Figure \ref{fig:pipeline_overview}. The implementation of each step can vary depending on the algorithm utilized. 
For example, other methods could be considered for generating entity embeddings, semi-supervised learning, article embeddings, NER, topic models, etc.
In this section, we present our implementation of the method and its main components.

\subsection{Articles and Entities Processing}
\label{sec_article_entity_processing}
Given a large corpus of articles, we extract and process the given articles into structured features, such as entities (persons, locations, and organizations), themes, and associated events. 
To accomplish this, we perform multiple steps: (a) removing redundant articles and rare entities, and handling article duplication, (b) applying source-based filtering, and (c) normalizing entities. In the following subsection, we will provide details on each step.

\subsubsection{Entities and Themes Extraction}
Given a corpus of textual articles $A$, we represent each article as a set of themes and entities. First, we suggest extracting those features from the textual corpus by deploying NER algorithms that extract locations, persons, dates, organizations, and the number of times each entity appears in a given article. Additionally, we utilize topic models to extract the topics and tones conveyed in the article.

\subsubsection{Processing Articles Entities}
The next step is to formalize the entities in each article with a global representation. 
Since multiple NER and topic models can be chosen, there might be some variance in how an entity is recognized.
We created mappers for each entity type that maps (Entity, Entity Name), e.g., (Person, Donald Trump), then we matched similar entities to their global representation, e.g., (Donald Trump, DONALD TRUMP).

We have implemented a naive method for performing entity matching. 
Namely, given the entity type, a set of normalizing operations was performed. We cast dates in a unified format.\footnote{We cast dates to the following formats: Day/Month/Year (dd/mm/yyyy) or  Day/Month (dd/mm).}
We cast persons and organizations into lower cases and locations to match their country and city codes. Next, we aggregated mappers for each entity type $<Entity, Count>$ during the pipeline. Lastly, when an entity that has not appeared yet is given as input, we check for the closest entity by measuring the hamming distance with a predefined threshold. Hamming distance is a popular measure of the distance or similarity between objects. The hamming distance is defined as the number of positions at which the corresponding symbols differ~\cite{bookstein2002generalized}. 
If the new entity does not have a match, it is inserted as a new entity to the relevant mapper.

\begin{figure*}[hbt!]
    \centering
    \includegraphics[width=0.7\linewidth]{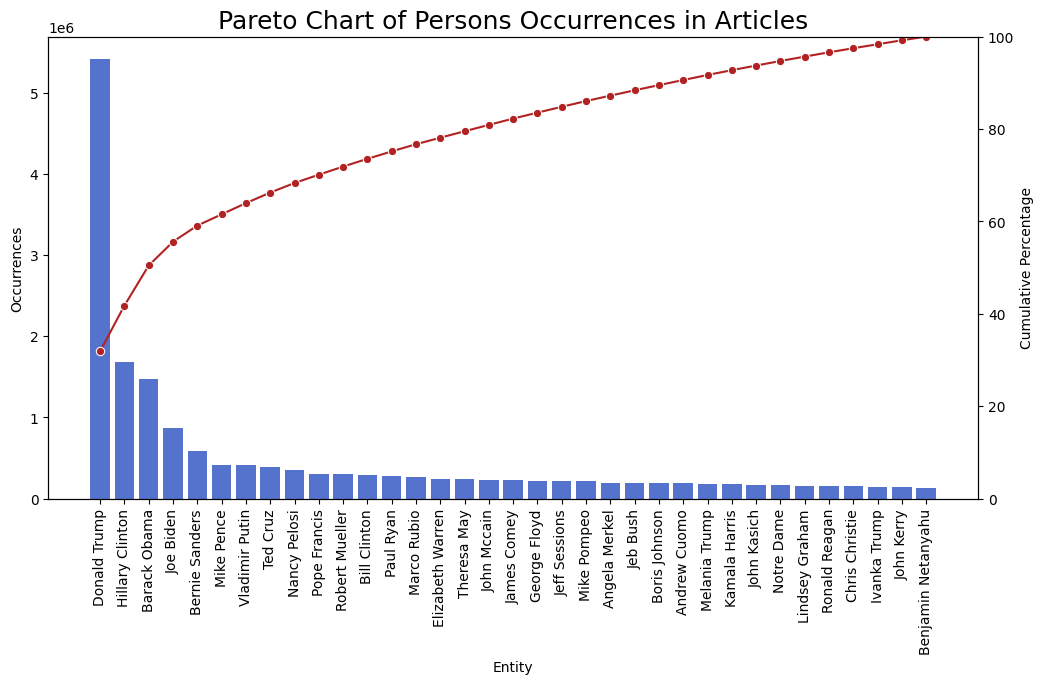}
    \caption[Pareto Chart of Person Occurrences in Articles]{This figure contained a Pareto chart of the Top 35 persons occurrences in the full collected dataset from GDELT.}
    \label{fig:pareto_pesons}
\end{figure*}

In many cases, large datasets of entities contain frequency distributed in a power-low manner (see, for example, Figure~\ref{fig:pareto_pesons}), meaning a few entities appear in most of the article's data. 
To optimize runtime and memory utilization, we handle the vast size of the entities by removing the entities that appeared less than the selected thresholds in the cleaned dataset.
We have manually defined the threshold parameters for each entity type based on the given data and the loss percentage of unique entities.

\subsubsection{Cleaning the Data}
After processing the entities and themes, the next step is to clean up redundant data. We identify two types of redundant articles: duplicates and very short articles. Article duplication can occur for several reasons, such as the same article being parsed multiple times from the same source or being obtained from different sources. To prevent over-representation of popular articles, we remove any articles that share the same themes, entities, and tones (provided by GCAM) as those already processed.
To handle the redundant short article, we removed all the articles containing less than a given predefined threshold of valid entities and themes.

\subsubsection{Splitting the Data}
Lastly, after completing the previous processing tasks, we implemented a splitting mechanism. This mechanism divides the given dataset of articles into smaller subsets based on a specified time frame. We do this for several reasons: first, the embedding representation of entities and articles should reflect the relevant contextual news.

For instance, we currently expect Joe Biden's embedding representation to be closer to the US President entity than Donald Trump's embedding. Additionally, to keep the method lightweight, we consider only part of the dataset at a time if necessary. Our study showed that most events have a short lifespan, so any information loss can be minimized by using a predefined time-frame split.

\subsection{Embedding Generation for Entities}
\label{subsec_entity_embed}
This stage is responsible for producing embedding representations for entities. Given time-separated datasets of processed entities found in articles, this stage generates embedding for each entity. 
To perform this, we train GloVe models for each time-partitioned dataset based on the entity co-occurrences in the articles. To improve the generation of time-sensitive embeddings for entities, we developed a pooling mechanism for additional articles before the training process to represent current and historical information.

\subsubsection{Embedding Generation for Entities - Adapting GloVe}
\label{sec_glove_adaptation}
To generate entity embeddings, we utilized the GloVe algorithm \cite{pennington2014glove}. 
We adopt the GloVe algorithm in the following manner: Instead of entering a list of words that appeared in a sentence, the GloVe model receives the set of entities and themes that appeared in the same article as input. We defined the context window size as the length of the largest set of entities in all the articles. Since we do not have the order of the entities that appeared in the articles, there is no meaning for a limited context window. 
The GloVe model incorporates global and nearby entities’ data and attempts to assimilate the benefits of the neural network language model. 

We use the following notations:\\
$X_{ij}$ is the number of times entity $i$ occurs in the same context window (article) as entity $j$. We define $X_i$ as: $X_i := \sum_j X_{ij}$, and $P_{ij} := \frac{X_{ij}}{X_i}$ is the probability that entity $j$ appears in the same article as the entity $i$. 
\newline
$W_i$ is a vector embedding representation of entity $i$.
For each entity, we define the following equation: 
\[log(X_{ij}) := w^T_i w_j + b_i + b_j,\]
where $b_i$ and $b_j$ are the biases of the central word and the context respectively. 

Overall, the loss function for GloVe preparation is displayed in the accompanying condition:
\[J = \sum_{i,j}f(X_{ij})(w^T_i w_j + b_i + b_j -log(X_{ij}))^2,\]
where $f(X_{ij})$ is a weight function.

Before training the GloVe model, we developed a pooling mechanism to ensure that the entities' embeddings generated for a specific period contain both recent and historical information. The pooling mechanism was constructed as follows: first, we sampled articles from previous periods in decreasing amounts, adding them to the current period corpus used to train the model. 
It is important to note that we did not sample data from the future to prevent leakage or bias in the entities' embeddings. The pooling mechanism is defined using the following notations: 
the dataset of all the processed articles' entities that appeared in a set time frame $t$ is marked as $M_t$.
We define $rand(M_k, p)$ to be a function that randomly selects articles' entities without replacement with $p$ probability to pick an article.
The article corpus for each time frame was assembled in the following manner:
\[Corpus_{M_k} := \{{M_k}, rand(M_{i-1}, p_1),..., rand(M_{n}, p_{n-1}).\}\]

\subsection{Embedding Generation for Articles}
\label{subsec_article_embed}
We utilized two distinct techniques to create news \textit{article embeddings}. 
Firstly, we deploy a sentence embedding model on the embedded entities space so that, given a set of embedding representations of article entities, this model generates a vector representing the article. 
Secondly, we trained a semi-supervised DNN, similar to Ohi et al. work \cite{OHI2020106190}; we did so on the tagged article event data. This model receives input from the naive article embedding to produce a new vector space that considers the cross-document relationships given shared events. 
After completing the two approaches, we generate a third and final representation for the article embeddings - where we rely on the synergy of the two previous methods.

In the following sections, we describe the steps required to generate our method’s final article embedding: deploy a sentence embedding model (see Section~\ref{sub_sif}), then train a DNN semi-supervised model (see Section~\ref{sub_siamese}). 
Lastly, we perform a synergy based on the previous steps (see Section~\ref{sub_concatenate}).

\subsubsection{Embedding Generation for Articles Using SIF}
\label{sub_sif}
In this stage, we construct a naive article embedding representations using the SIF algorithm \cite{arora2017simple}.
Conceptually, to represent a complex object, one can apply a simple aggregator function on the object's component embeddings. 
By this logic, the SIF model receives as input a list of the entities' embeddings that appeared in each article, as generated in the previous step (see Section~\ref{subsec_entity_embed}), and outputs article embeddings.

We chose the SIF model as the aggregator function~\cite{arora2017simple} due to its simplicity and competitive performance. 
The underlying intuition is that the embeddings of too frequent words should be down-weighted when summed with those of less frequent ones, which is compatible with the use-case of entities in articles - frequent entities provide less context required for distinguishing articles from one another. 

We implemented this and considered it as the base approach for the task of article embeddings.
However, there might be a drawback to such an approach, it might suffer from underfitting. 
The simplicity of the SIF model does not account for additional information given by a group of certain entities.
To uncover that pattern, one can use the information between articles - in our case articles that discuss common events.

\begin{figure*}[hbt!]
    \centering
    \vspace{0.25cm}
    \includegraphics[width=0.7\linewidth]{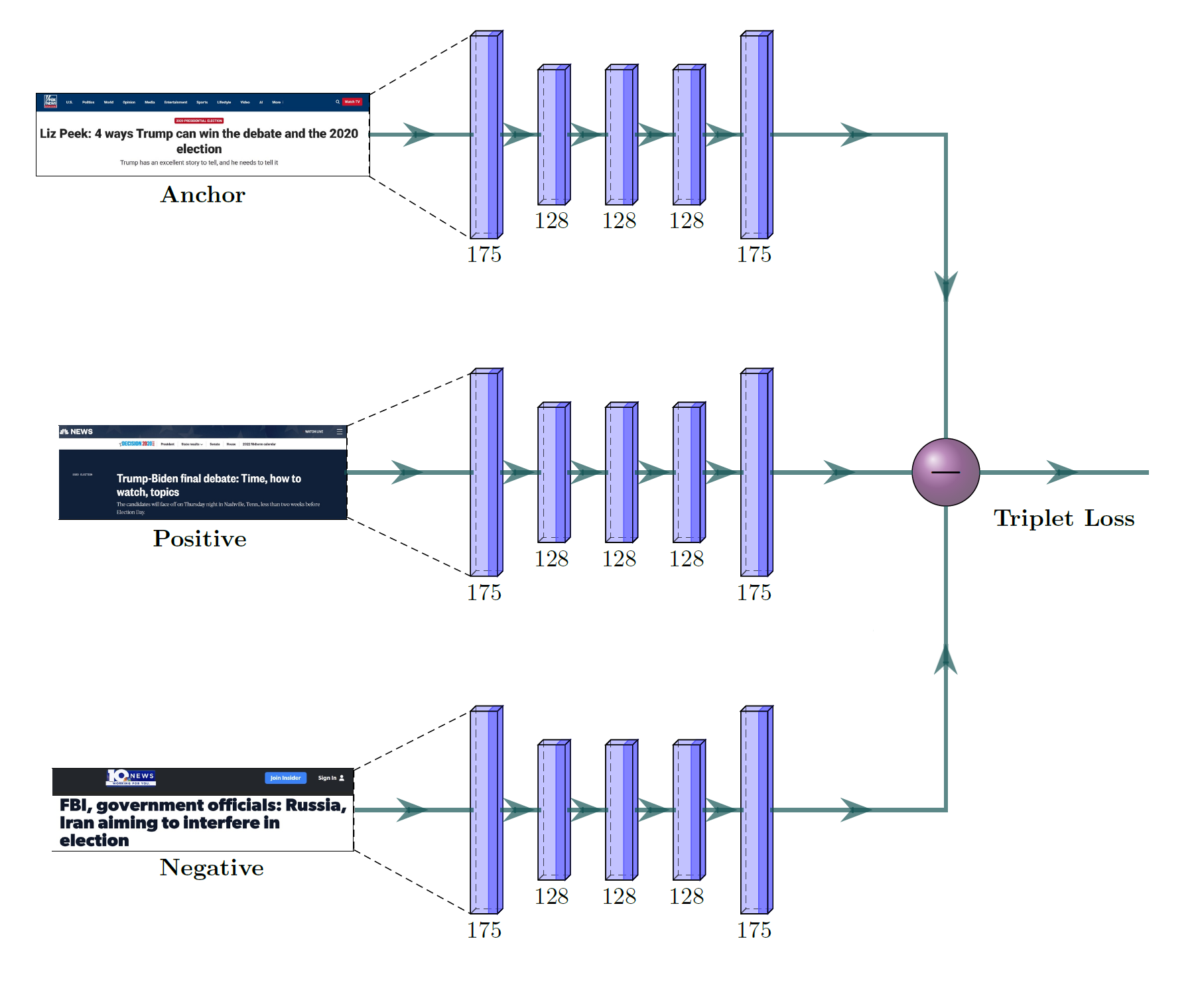}
    \caption[Triplet Siamese Network Training Mechanism]{The architecture and training process of each Triplet Siamese Network.}
    \label{fig:siamese_net_overview}
\end{figure*}

\subsubsection{Embedding Generation for Articles Using DNN Semi Supervised}
\label{sub_siamese}
We chose the Siamese Network model for this task, introduced by Bromley et al. \cite{bromley1993signature}, based on two identical DNNs.
The central concept of the approach is independent of the complexity of the model. 
Therefore, we chose a basic architecture for the models, which is independent of the specific time frame of the training data. 
This means that we applied the same model architecture to the entire dataset.
The input layer of the model equals the size of the embedding generated by the SIF model.
Three hidden layers are densely connected.
The output layer of the model equals the input embedding size.
The ReLU activation functions were configured on top of each layer, with uniform kernel initialize functions and L2 regularization functions. 
Figure \ref{fig:siamese_net_overview} presents an overview of the suggested implementation of the Siamese Network.

Similar to Ren et al.~\cite{ren2020intention}, we have also trained a Siamese Network model, with a Triplet Loss objective. 
Therefore, in our implementation, the architecture of the Siamese model contained three identical DNNs, each of which was built based on the architecture properties we described above.
Triplet Loss objective is a distance-based loss function that operates on three inputs:
\begin{enumerate}
    \item anchor (a) is any arbitrary data point.
    \item positive (p) which is the same class as the anchor.
    \item negative (n) which is a different class from the anchor.
\end{enumerate}
Namely, it is defined as: 
\[L=\max(dist(a,p)-dist(a,n)+\epsilon, 0).\]
Our goal in using this network architecture is to generate article embeddings. For our purposes, we define the anchor as an article. An article that shares common events with the anchor is considered a positive sample, while an article that does not is considered a negative sample.

We deployed the following triplet mining policy: To construct a triplet for a particular anchor article $a$, we must select a positive article, $p$, that shares at least one mentioned event,
and a negative article, $n$, that does not share any common event with $a$. In a dataset with $N$ training articles, there are $O(N^3)$ possible triplets,
many of which do not help the training converge (e.g., triplets where $dist(a,n) >> dist(a,p))$.

Many works have discussed the benefit of hard negative mining in constructing triplets that produce useful gradients and therefore help triplet loss networks converge quickly \cite{schroff2015facenet, simo2015discriminative}. 
However, previous works also show that hard positive examples can increase clustering within a class \cite{hermans2017defense}. 
Therefore, we defined the following policy using the ADAM optimizer~\cite{kingma2014adam}: we first sample $N$ triplet pairs, and out of those, we select the top $K_1$ most violated matches in each sample. 
This is done by computing the pairwise distances between all articles within the initial batch. 
The pairwise distances were calculated as follows:
\[dist(a,p) - dist(a,n) = \sum(a-p)^2 - \sum(a-n)^2.\]
To avoid clustering within a specific class, in addition to the $K_1$ most violated matches, we added to the batch $K_2$ random matches from the initial sample. 

As suggested by Roy et al.~\cite{roy2018action}, to generate the article embedding using the trained Siamese model, we consider a single DNN from the Siamese network triplet and utilize it to produce the article embeddings.
We do so by taking the output of the last layer in the selected DNN, that is the same size as the input layer.

\subsubsection{Embedding Generation for Articles - Synergy}
\label{sub_concatenate}
After the completion of the training process of the Siamese network, we generated two different embedding representations for all articles given their entities: 
(a) the naive SIF approach, and (b) the data-driven Siamese network approach.

This stage relied on the premise that there is synergy between those approaches; each approach has its merit, and depending on the data, each might contribute additional information. 
To consider this, in the final stage, the algorithm receives as input those embedding representations and outputs a vector of their concatenation, which represents an event-centric embedding of the given article.

\subsubsection{Articles Embedding Evaluation}
As explained in the previous sections, each method generated an embedding representation for each article by receiving the article's entity embedding as input.
We perform extensive experiments to evaluate the performance of the proposed methods for event-centric article embedding. 
We measure the methods' performances on the following task: Given a periodic dataset of articles' embeddings and their tagged events, we create pairwise permutations of all the articles in the given dataset. 
A pair of articles is considered close or instead tagged as positive if both articles share at least a single joint event. 
A simple Euclidean distance between each pair's embedding representation concludes their ranking score.

To validate the performance of our method, we performed a cross-dataset evaluation on different periods across the collected data without fine-tuning the models with the variant training data.  
The evaluation process considered the three proposed methods for article embeddings. 
To measure the methods, we consider the task of common event attribution and evaluate two metrics: precision-recall AUC and ROC AUC \cite{davis2006relationship}.
Moreover, we also conducted the Friedman test \cite{friedman1937use} and Nemenyi post-hoc \cite{nemenyi1963distribution} to determine whether a significant statistical difference exists between the article embedding generation methods.


\section{Experimental Setup}
\label{sec_experiments}
We evaluated our methods (see Section \ref{sec_method} and Figure \ref{fig:pipeline_overview}) on monitored news sources by GDELT from 02/2015 to 01/2021. 
Finally, after the embedding generation of articles is complete, cosine similarity between all the article pairs in the same time frame is computed to determine which articles share at least a single common event. 
The dataset is described in detail in Section~\ref{sec_data}, the experiments are described in Section~\ref{sec_experimental_setup}, and the results are presented in Section~\ref{sec_results}.

\subsection{Data Description}
\label{sec_data}
Throughout this paper, three different types of information are employed, including global knowledge graphs, events, and article-event mentions. 
These datasets are accessible by GDELT as three different types of files, for additional information on the GDLET source data (see Appendix~\ref{sec_additional_data_notes}). 
The period of the data that have been collected is 02/2015 to 01/2021.
Overall, in this research, we have evaluated 850,492 unique articles, 1,002,967 events, and 3,551,906 mentions.

We decided to collect all the articles from various popular media sources; each is viewed significantly more by people who identify as liberals or conservatives. We depended on previous works that classified different sources to a specific political view, such as described in a study by Iyengar et al.~\cite{iyengar2009red}. 
We have also decided to monitor the affiliates of those networks if they exist. 
For example, Fox News has more than 50 local affiliates, so we collected data on the articles from all of Fox News affiliates and classified them as the same source 'FoxFamily'. 
Including affiliates, we have gathered articles from 605 media sources (see Appendix~\ref{sec_dataset_description}).
Figure \ref{fig:sources_volume} presents the distribution of media sources in the collected data sets, labeled by political affiliation.

\begin{figure*}[hbt!]
    \centering
    \includegraphics[width=0.7\linewidth]{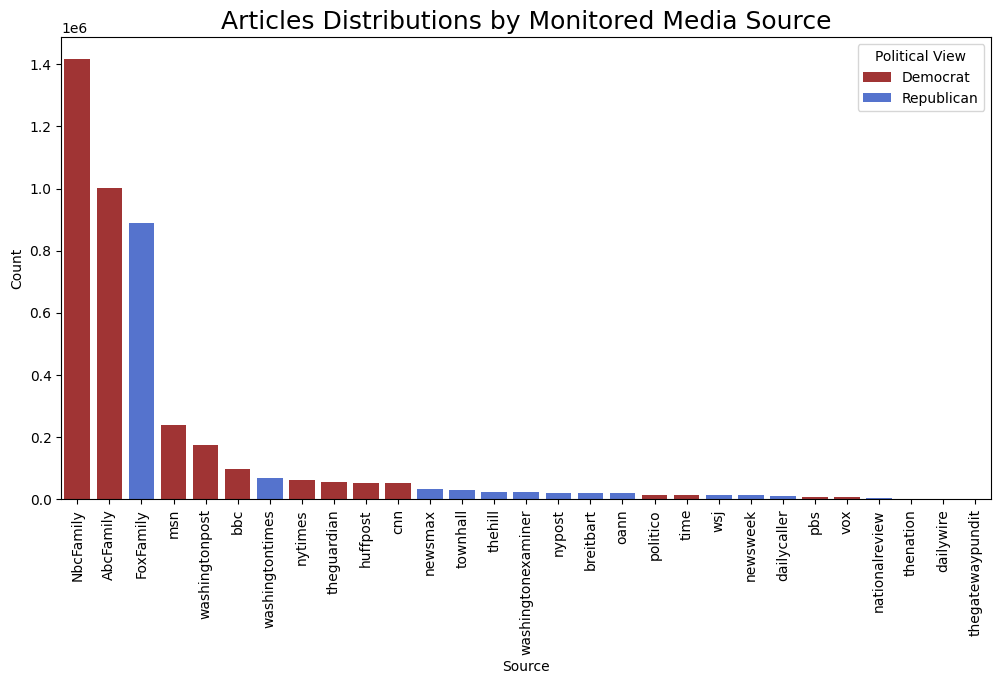}
    \caption[Media Source Article Mapping Histogram]{This figure maps each monitored media source to the articles published by it in the preprocessed dataset. 
    Conservative and Liberal media sources are labeled in blue and red, respectively.} 
    \label{fig:sources_volume}
\end{figure*}

\begin{table*}[hbt]
    
    \caption[Yearly Aggregated Datasets Description]{Yearly aggregated datasets statistics used in this work for performance analysis.}
    \renewcommand{\arraystretch}{1.25}
    \centering
    \vspace*{3mm}
    \resizebox{11cm}{!}{
    \begin{tabular}{c|c|c|c|c|c}
        Year & \#Articles & \#Events & \#Mentions & \%Monthly PR & \%Daily PR \\ \hline
        2015 & 78877 & 96783 & 361148 & 0.0547 & 0.5012 \\ 
        2016 & 166754 & 196716 & 702328 & 0.0423 & 0.3630 \\ 
        2017 & 162264 & 199470 & 694312 & 0.0436 & 0.3764 \\ 
        2018 & 160505 & 192327 & 681872 & 0.0470 & 0.4066 \\ 
        2019 & 139823 & 157397 & 555292 & 0.0398 & 0.3515 \\ 
        2020 & 142269 & 160274 & 556954 & 0.0381 & 0.3253 \\ 
        \hline
        \textbf{-} & \textbf{850,492} & \textbf{1,002,967} & \textbf{3,551,906} & \textbf{0.043\%} & \textbf{0.377\%} 
    \end{tabular}
    }
    \label{tab:table_yearly_aggregated_stats}
\end{table*}

\subsubsection{Labeled Datasets Construction}
We validate the proposed methods by performing cross-dataset testing by using our method. We have separated the entire collected article data into 66 datasets, separated by months. 
We derived from the GDELT open-sourced data, and Table \ref{tab:datasets_statistics} presents the characteristics of each monthly dataset included in the experiment.
Table \ref{tab:table_yearly_aggregated_stats} shows yearly aggregated statistical information of those datasets.
It is important to note that there is a variance in the datasets' size. 
The difference in the datasets' sizes led us to assume there was an inconsistency in the raw GDELT data. 

We evaluated our method using these datasets on two tasks: monthly and daily attribution of two articles to share events.
\begin{itemize}
  \item \textbf{Monthly Mentioned Events}: 
      A large-scale dataset consists of every possible permutation of 2 articles in the same month.  Given the embedding representation of each pair of articles, we calculate a prediction probability score that those articles shared at least a single joint event.  It is worth noting that working with this dataset is challenging because it is highly imbalanced, and the probability of pairing articles with relatively distant creation dates is substantially lower since events' mentions appearances over time were skewed toward their creation date.
    \item \textbf{Daily Mentioned Events}: 
      A smaller-scale dataset consists of every possible permutation of 2 articles published on the same day. 
      Each pair of articles is assigned a probability score for that pair of articles to share a common event. 
      It is worth noting that this dataset is less challenging because it is considerably more balanced than the monthly use case.
\end{itemize}

\subsection{Experiments}
\label{sec_experimental_setup}
In this section, we will outline the implementation details and discuss the evaluation process for each step of our proposed methodology. We conducted experiments to determine whether our approach improves the performance of common event attribution in news articles, whether this improvement is consistent across different data distributions, and whether our approach is independent of the initial embedding method. 
For additional implementation notes regarding the research experiments, see Appendix \ref{sec_deep_implementation_notes}.
\\\\
\textbf{Articles and Entities Processing.}
First, as we mentioned in Section \ref{gdelt_subsection}, GDELT publishes the output of its methods to extract Persons, Locations, Organizations, Dates, Themes, and Tones. So, we extract those from the GDELT raw data and construct entries that contain the article: URL, entities, themes, events, time, and tones.

In the processing, we defined the following parameters: 
\begin{itemize}
    \item The entities' pruning thresholds were set as (Themes, 20), (Persons, 75), (Organizations, 50), (Dates, 25). Locations entities were not removed, as they contain relatively a much smaller set of unique values. Those parameters were manually set with the condition of keeping the most common 33\% of the entities for each entity type.

    \item For determining whether an entity already appeared, the hamming distance similarity threshold was set to two. We have picked the threshold by manually selecting a few entities and searching for their closest entities by hamming distance. 
    We noticed that, in the majority of cases, a hamming distance of two or less was caused by a typographical error.

    \item To handle the redundant short articles, we removed all the articles containing less than four valid entities and themes. 
    We assume that such article is irrelevant or that there was an error in its entity extraction process.

\end{itemize}

{\begin{figure*}[hbt!]
\centering
\includegraphics[width=0.7\linewidth]{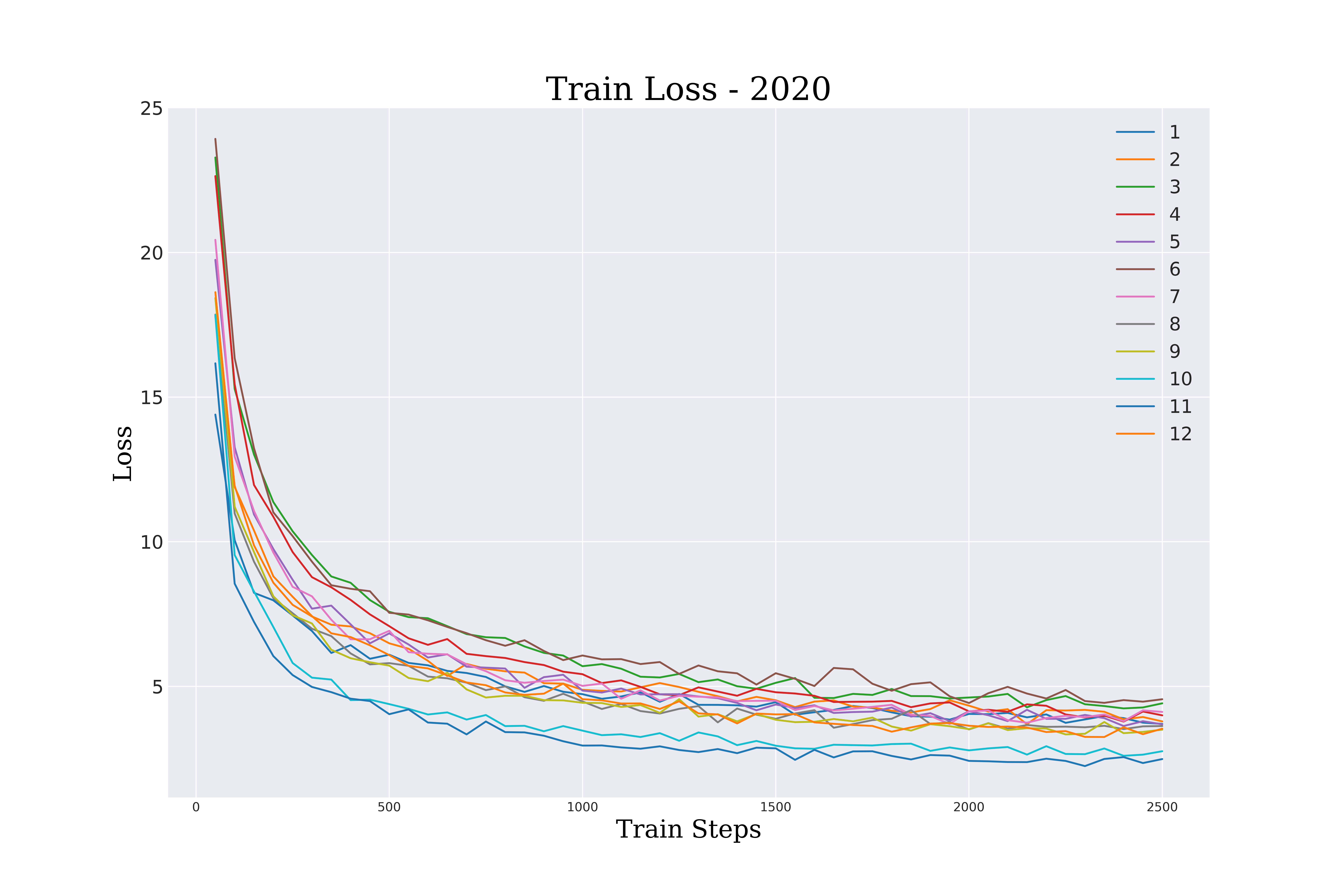}
\caption[Train Loss Figure 2020 Models]{Siamese-Network: Train Triplet Loss. The X-axis is the monitored training steps; for every 4 steps, the average loss was calculated. The labels represent each Siamese model that was trained for a given month.}
\label{fig:siamese_net_train_loss}
\end{figure*}}

\noindent \textbf{Embedding Generation for Entities.}
We decided to collect a corpus of the recent six months, using the pooling mechanism we defined in Section \ref{sec_glove_adaptation}. 

The GloVe parameters were set as
The GloVe parameters were set as learning rate = 0.05, max loss = 10, and vector size = 175.
We fixed these parameters for all the experiments.

We manually evaluated the generated GloVe embedding by doing the following: We selected 50 different themes and entities overall. 
For each entity or theme, we searched for the five closest neighbors on a monthly trained GloVe model. We had two criteria for selecting each entity:

(a) The GloVe model successfully accounted for recent information regarding that entity. For instance, we checked the closest entities to Nancy Pelosi's embeddings during January 2021, the month of the storming of the Capitol, in which protesters took over Nancy Pelosi's office. We checked if the embedding of the 'White House' was a close neighbor.

(b) The embedding of that entity accounted for historical information. For example, we used the entity '9.11' and checked if we found the theme 'TERROR' in its close neighbors.

After manually examining the selected entities and themes, we found their relationships with others to be reasonable. For examples of the pooling mechanism's impact on the entities' embeddings, see Appendix~\ref{sec_Embedding_Generation_Notes}.
\\\\
\textbf{Embedding Generation for Articles.} 
In each experiment, we split the events in the given dataset by a 70\%-30\% ssplit to train and test by their order of appearance, as stated by Nguyen et al. ”The 70–30 split, widely employed in machine learning, strikes a balance between an adequate training set size and a sufficiently large test set, ensuring a robust evaluation of the model’s performance” \cite{nguyen2023utilization}.
In addition, data leakage articles duplication between the train and test set were removed.

When training the Siamese networks, we defined fixed values for the hyperparameters: ADAM optimizer with a learning rate of 0.0015. 
The Siamese Triplet model contained three DNN network architectures: simple input and output layers of 175 units, with three hidden layers of 128 units. We have also defined L2 regularization on top of each layer of the model. 
Each batch had a size of 64, which consisted of 50 random pairings of articles and 14 hard pairings, as described by our defined triplet mining policy.\footnote{The models were not trained on epochs but on 10,000 batches because of the sheer volume of the pair article's permutations.}
We chose 10,000 batches since we saw convergence across the datasets beforehand, which made sense in the relatively small network we constructed. This can be exemplified by the train loss of the Siamese models we train in 2020 monthly datasets (see Figure~\ref{fig:siamese_net_train_loss}).
Each label is the relevant Siamese model for a given month. 
We constructed the embeddings, as defined in \ref{sub_siamese}.
\\\\
\textbf{Performance Evaluation}
To evaluate the performance of our approach to generating news embedding, we performed the following steps:
\begin{enumerate}
    
    \item We deployed the semi-supervised event approach, SIF, and their concatenation – for all tasks of common event attribution.
    \item We evaluated their performance on those tasks on all the collected datasets from GDELT, to determine whether the improvement was generalized over different data distributions.
    \item We implemented a spaCy language model to generate embedding for articles given their themes and entities and evaluated them against our method. 
    \item Lastly, we deployed our semi-supervised approach on the spaCy models’ embeddings to verify that the approach is agnostic to the initial embedding method. This Siamese network was trained in the same manner as we described above.
    
\end{enumerate}

\section{Results}
\label{sec_results}
In this section, we present the results of the experiments as described in the preceding Section \ref{sec_experimental_setup}.

\subsection{Periodical Theme and Entity Embeddings}
In this section, we present the number and type of each periodic embedding we generated in this work. As part of our study’s methodology, we have developed a new way of generating embeddings that are relevant to different periods. Specifically, we have created monthly embeddings for every valid entity or theme that we have identified, as defined in Section \ref{sec_article_entity_processing}.
This means that even if a valid entity appears only once in all the monitored media sources in a given month, we can still create a relevant monthly embedding representation for it by using the GloVe model training process, as we described in Section \ref{subsec_entity_embed}.
Our methodology involved monitoring various features of the news articles, including Person, Location, Date, Organization, and Theme. 
We constructed monthly embeddings for distinct entities from 02/2015 to 01/2021, and the count of these entities is presented in Table \ref{tab:monitor_entity_type_counts}. 
Throughout the 66 months of monitoring, we aimed to generate 331,630 embeddings for news entities and themes each month.

{\begin{table}
\centering
\caption[Monitored Entity Type Count]{The number of unique generated embeddings per monitored entity type.}
\vspace{0.25cm}
\renewcommand{\arraystretch}{1.25}
\label{tab:monitor_entity_type_counts}
\begin{tabular}{c|c}
\centering
Entity Type  & Count   \\ \hline
Person       & 202,006 \\
Organization & 103,201 \\
Theme        & 16,577  \\
Date         & 9,552   \\
Location     & 294     \\
\hline
\textbf{Total}            & \textbf{331,630}         
\end{tabular}
\end{table}
}

\begin{table*}[hb]
\centering
\caption[Summarized Methods Performance Results]{ The mean and standard deviation, of each method on both daily and monthly common event attribution tasks, for PR and ROC AUC metrics.}
\vspace{0.5cm}
\renewcommand{\arraystretch}{1.25}
\adjustbox{max width=\textwidth}{%
\centering
\begin{tabular}{l|cccc}
\hline
\multicolumn{1}{c|}{\multirow{2}{*}{Method}} & \multicolumn{2}{c}{Monthly Evaluation}        & \multicolumn{2}{c}{Daily Evaluation}          \\
\multicolumn{1}{c|}{}                        & Precision-Recall AUC  & ROC AUC               & Precision-Recall AUC  & ROC AUC               \\ \hline
SIF                                          & $0.348 \pm 0.048$          & $0.958 \pm 0.015$          & $0.547 \pm 0.055$          & $0.955 \pm 0.014$          \\
Siamese Network                              & $0.344 \pm 0.054$          & $\mathbf{0.974 \pm 0.012}$ & $0.548 \pm 0.060$          & $\mathbf{0.972 \pm 0012}$  \\
Concatenation                                & $\mathbf{0.369 \pm 0.048}$ & $\mathbf{0.974 \pm 0.010}$ & $\mathbf{0.572 \pm 0.055}$ & $\mathbf{0.972 \pm 0.010}$ \\ \hline
spaCy CPU Language Model                                           & $0.009 \pm 0.003$          & $0.877 \pm 0.025$          & $0.068 \pm 0.020$          & $0.876 \pm 0.025$          \\
Siamese Network - spaCy CPU Language Model                                           & $0.012 \pm 0.004$          & $0.933 \pm 0.018$          & $0.089 \pm 0.028$          & $0.932 \pm 0.018$          \\
\hline
\end{tabular}
}
\label{tab:summarized_reults}
\end{table*}

{\begin{figure*}[ht!]
\centering
\subfloat[Monthly Precision-Recall AUC Comparison]{\includegraphics[width=0.3\textwidth]{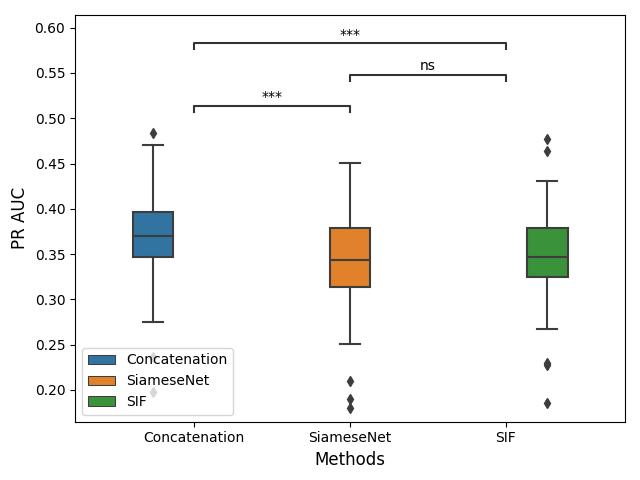}}
\subfloat[Monthly ROC AUC Comparison]{\includegraphics[width=0.3\textwidth]{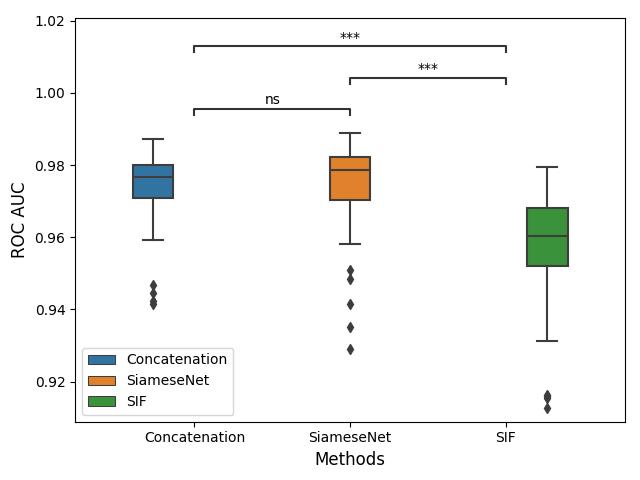}}\\
\subfloat[Daily Precision-Recall AUC Comparison]{\includegraphics[width=0.3\textwidth]{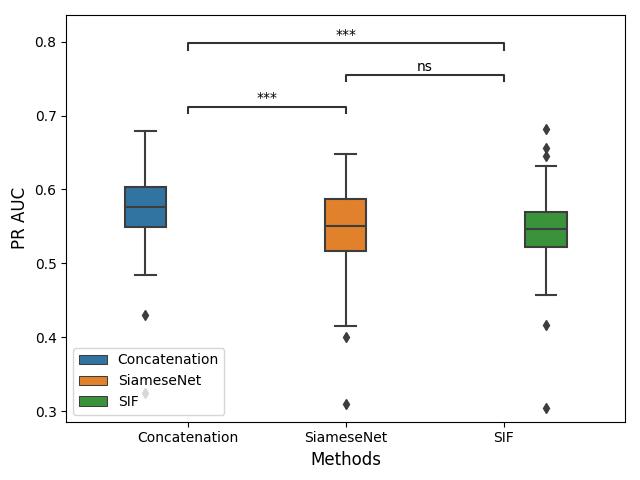}}
\subfloat[Daily ROC AUC Comparison]{\includegraphics[width=0.3\textwidth]{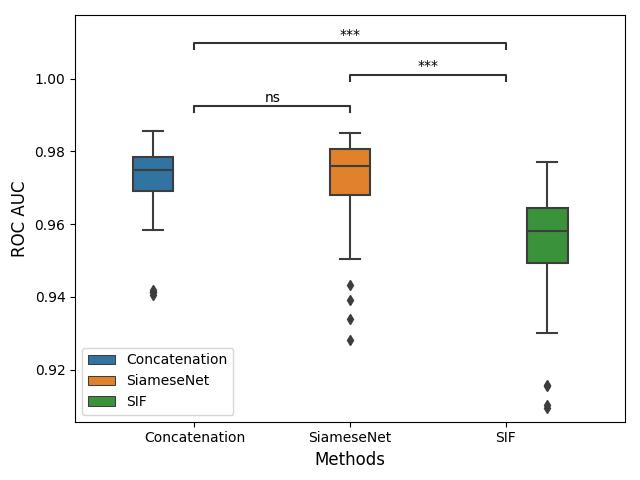}}
\vspace{0.25cm} 
\caption{Statistical Analysis of Article Generation Methods - Comparison of performance using the Friedman and Nemenyi tests with a 5\% significance level in the daily and monthly event attribution task.}
\label{fig:box_plot_methods_comparison}
\end{figure*}
}

{\begin{figure*}[ht!]
\centering
\subfloat[Monthly Precision-Recall AUC Comparison]{\includegraphics[width=0.3\textwidth]{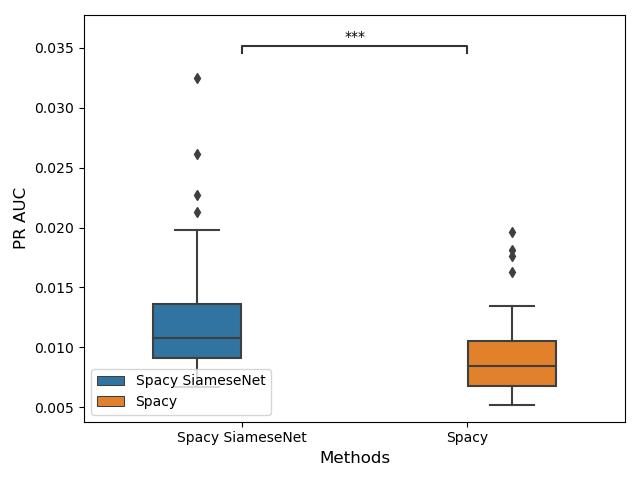}}
\subfloat[Monthly ROC AUC Comparison]{\includegraphics[width=0.3\textwidth]{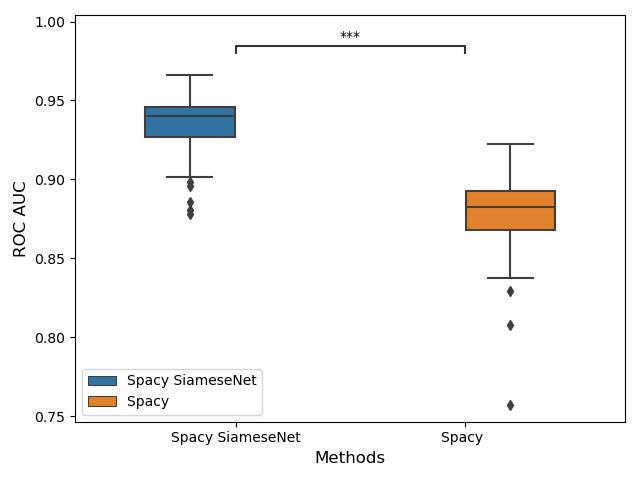}}\\
\subfloat[Daily Precision-Recall AUC Comparison]{\includegraphics[width=0.3\textwidth]{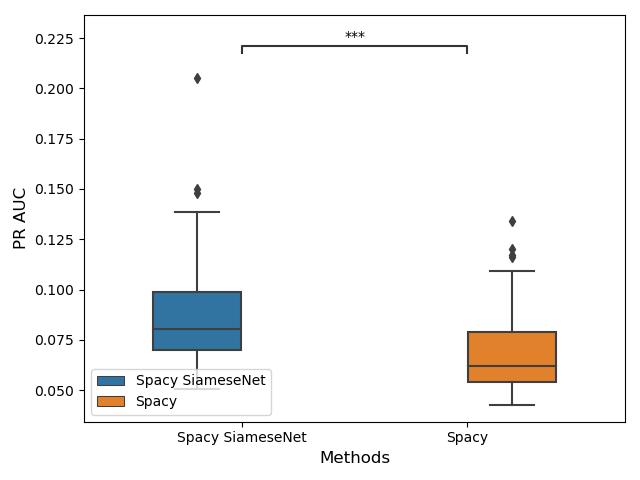}}
\subfloat[Daily ROC AUC Comparison]{\includegraphics[width=0.3\textwidth]{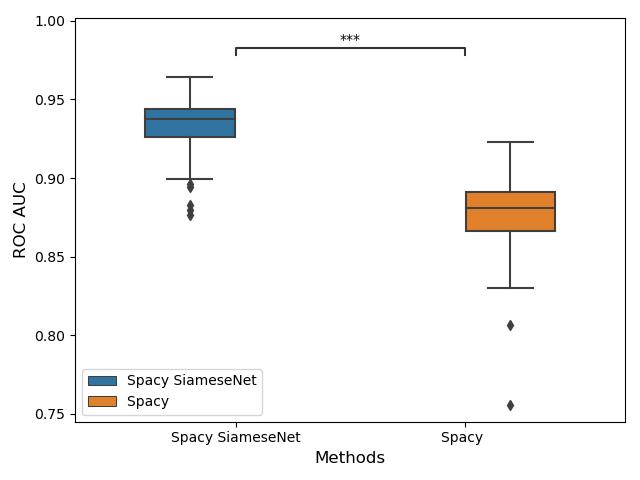}}
\vspace{0.25cm} 
\caption{Statistical Analysis of Article Generation Using the spaCy Model - Comparison of performance using the Friedman and Nemenyi tests with a 5\% significance level in the daily and monthly event attribution task.}
\label{fig:box_plot_spacy_comparison}
\end{figure*}
}

\subsection{Common Event Attribution Tasks}
As described in Section \ref{sec_experiments}, we ran the experiments on all 66 distinct monthly datasets we collected and processed from the GDELT project.
The results of those experiments on our suggested approach which considered both events and the time-based entity embeddings appear fully in Figure \ref{fig:total_methods_comparison}.

To summarize these results, we also constructed the following Table \ref{tab:summarized_reults}, which contains the average performance of all methods, across all datasets, measured metrics, and tasks.

Additionally, in Figures \ref{fig:box_plot_methods_comparison}, we visualize the distribution of the performance of all methods in the context of metrics (ROC and PR AUC) and tasks, both monthly and daily, across all the datasets. 
Moreover, we performed statistical analysis to determine whether the methods had a significant difference. We concluded from the Precision-Recall AUC evaluation that the concatenated approach provides better performance in both tasks compared to the standalone baseline and Siamese network.

\subsection{Comparison to Existing Method}
The same experiments were performed on the spaCy large CPU model, the results for those experiments are shown fully in Figure \ref{fig:spacy_methods_comparison}.
These experiments were performed to showcase:
\begin{enumerate}[label=(\alph*)]
    \item Comparing our methodology to out-of-the-box solutions for article embeddings.
    \item Evaluating if our semi-supervised approach yields an agnostic improvement to the generated articles embedding, relative to the given basic article embedding representation.
\end{enumerate}
To perform this comparison, we also included in Table \ref{tab:summarized_reults} the results of the basic spaCy model and the semi-supervised infused spaCy model.
Additionally, to determine the significant improvement of our approach, we included Figure\ref{fig:box_plot_spacy_comparison}, which evaluated the performance of each model embedding for all the experiments of common event attribution.

{\begin{figure*}[ht!]
\centering
\subfloat[Precision-Recall AUC Comparison]{\includegraphics[width=0.5\linewidth]{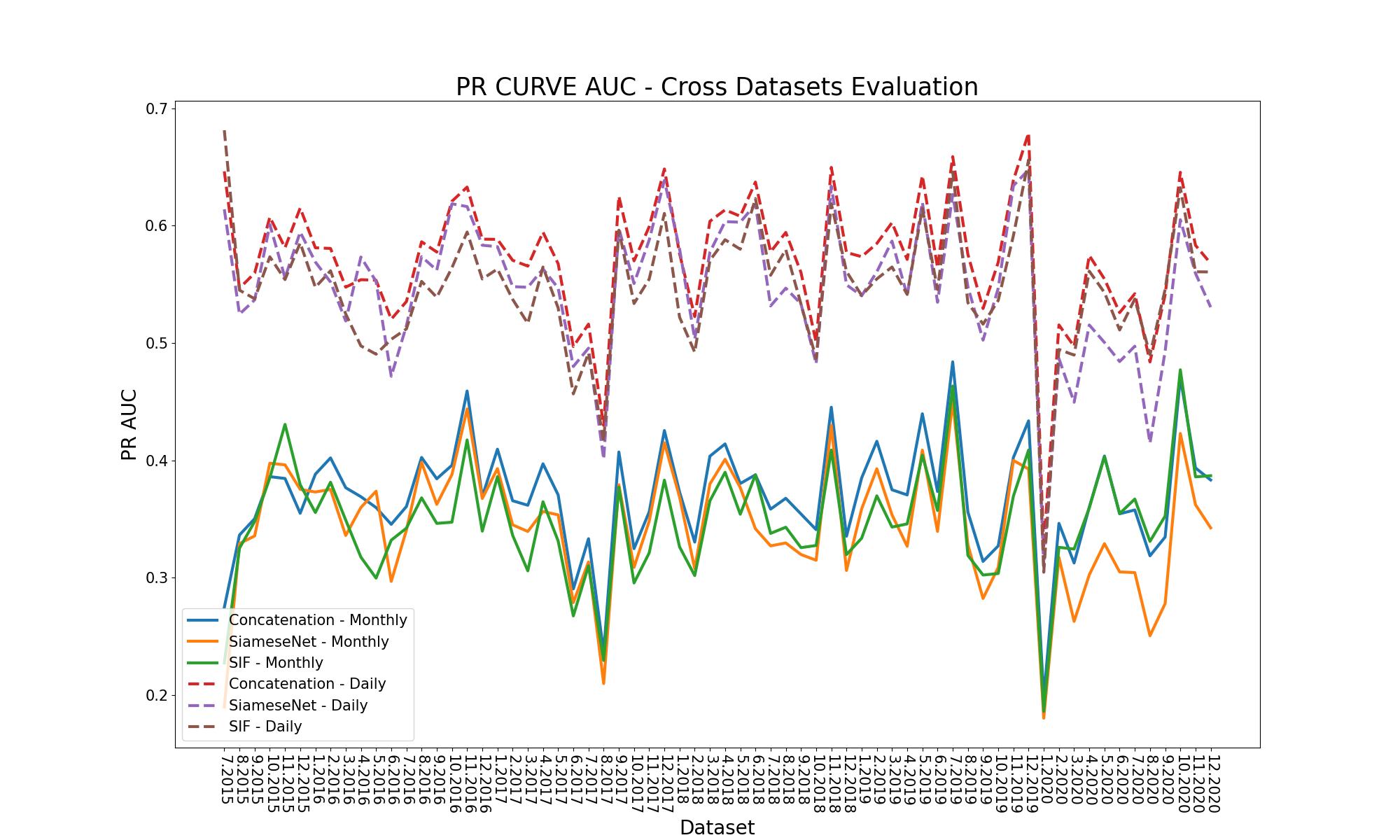}}
\subfloat[ROC AUC Comparison]{\includegraphics[width=0.5\linewidth]{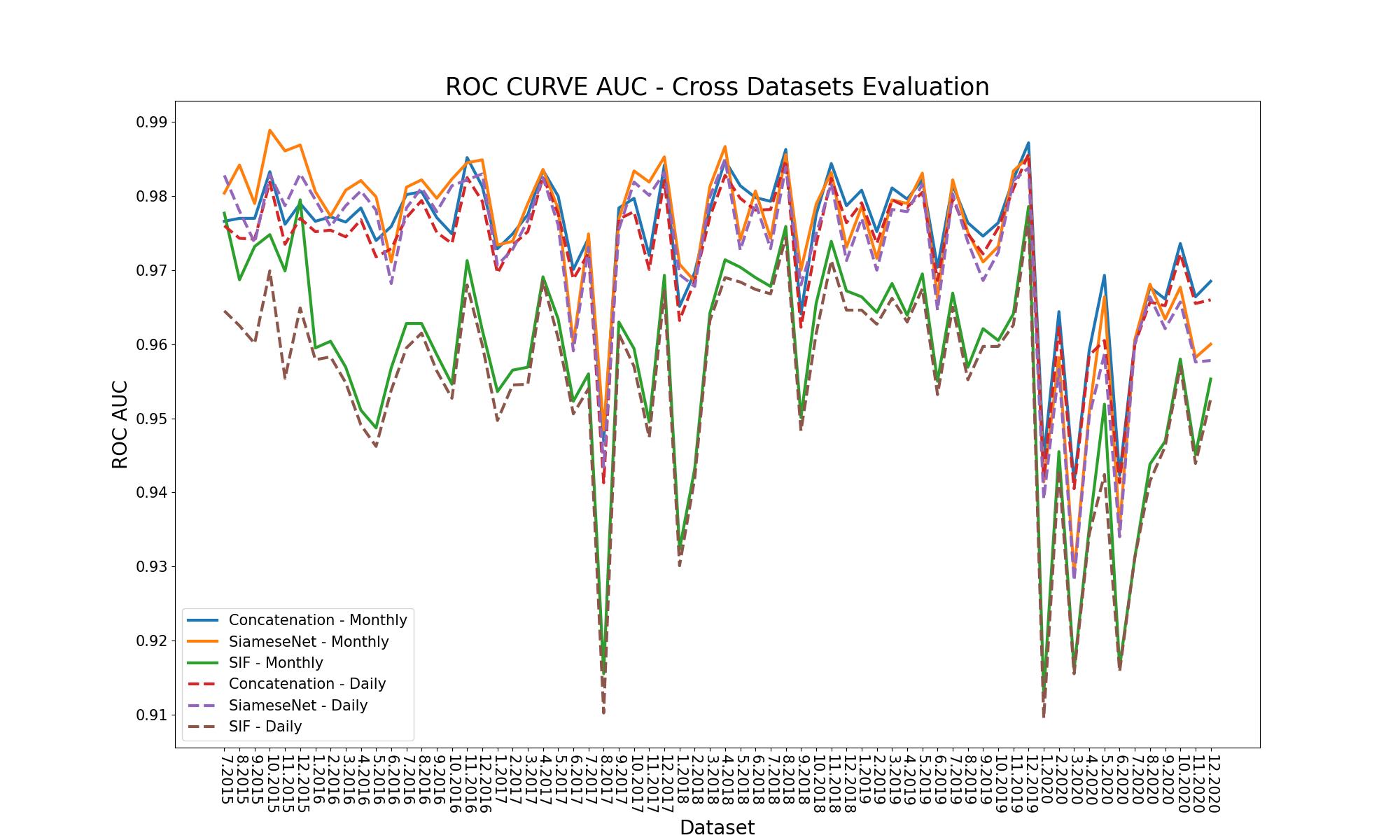}}
\vspace{0.25cm} 
\caption{ROC and Precision-Recall AUC results of the daily and monthly common event attribution task across all methods and datasets.}
\label{fig:total_methods_comparison}
\end{figure*}}

{\begin{figure*}[ht!]
\subfloat[Precision-Recall AUC Comparison]{\includegraphics[width=0.5\linewidth]{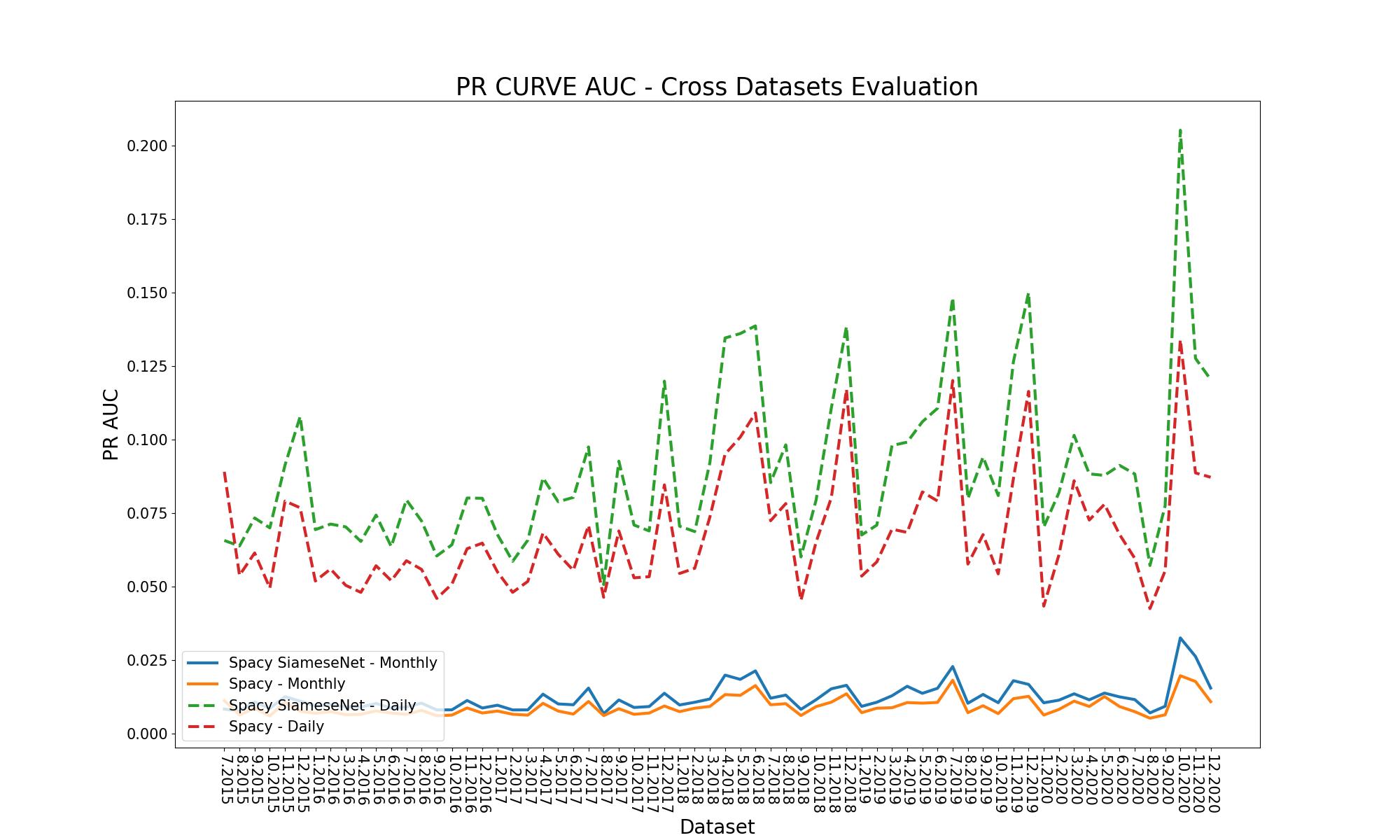}}
\subfloat[ROC AUC Comparison]{\includegraphics[width=0.5\linewidth]{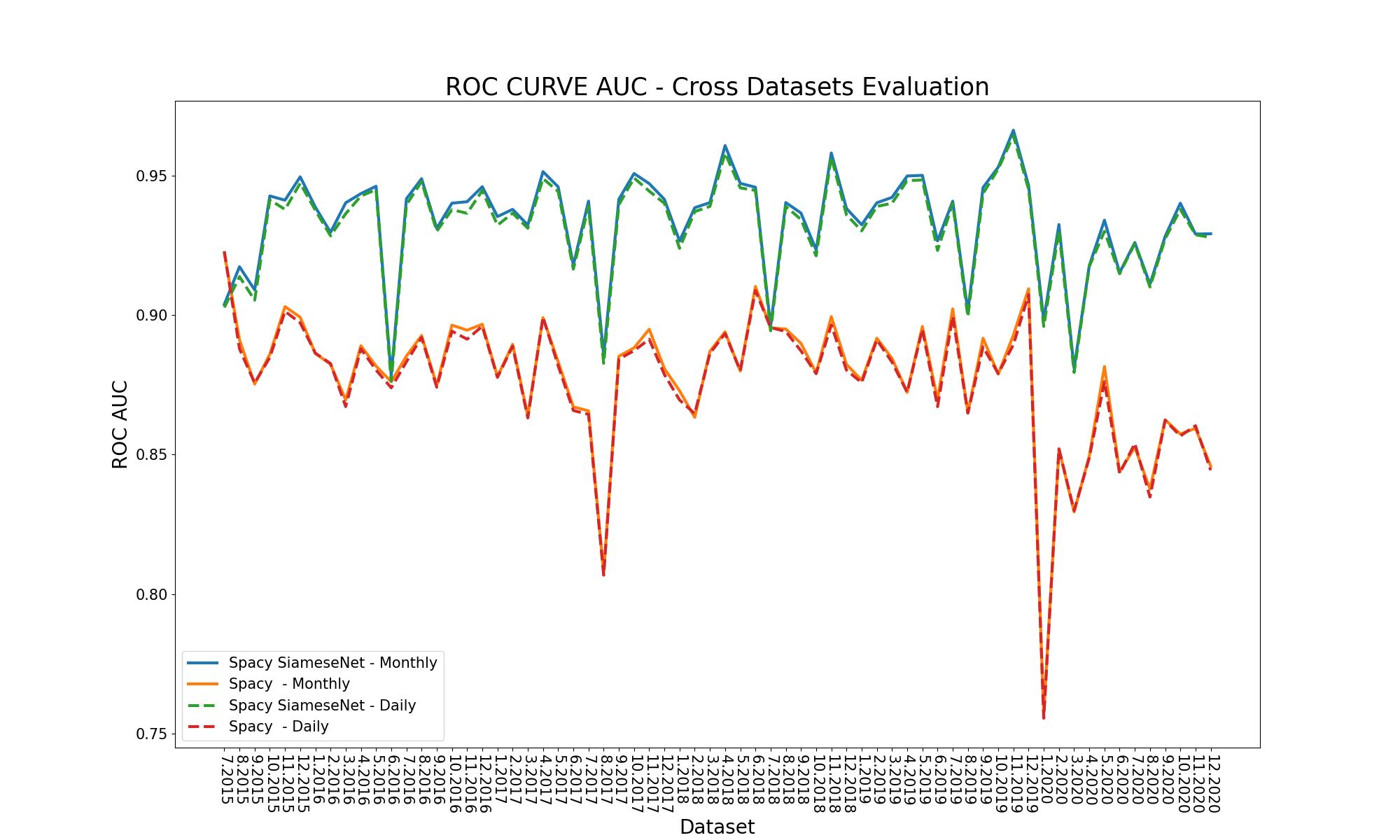}}
\vspace{0.25cm} 
\caption{ROC and Precision-Recall AUC results of the daily and monthly common event attribution task across all methods and datasets.}
\label{fig:spacy_methods_comparison}
\end{figure*}}

\subsection{Statistical significance evaluation}
\label{section:statistical_tests}
We perform statistical analysis to verify the existence of significant differences among the evaluated methods. 
As suggested by \cite{demvsar2006statistical}, we use the Friedman test \cite{friedman1937use} on the results of the aforementioned approaches to identify which obtained better results regarding each designed task and metric across all datasets.

The tests were used with a significance level of 5\%.
Specifically, the rank results established by the statistical method provide the criterion for evaluating the approaches, considering that the higher the value of the calculated rank, the better the result will be in this approach. 
In our case, the null hypothesis was rejected in each context, as can be seen in Table \ref{tab:friedman_results}.
The rejection of the null hypothesis of the Friedman test indicates the existence of statistical differences between the methods but does not define which of them are statistically different. 
For that, the Nemenyi post-hoc test \cite{nemenyi1963distribution} was applied to make multiple comparisons using the values of the percentages of means for each approach evaluated in all the executed datasets, as can be seen in Table \ref{tab:Nemenyi_results}.
The results of the statistical analysis are presented in Appendix~\ref{appendix_statistical_test_results}.

Therefore, in all cases, there were significant differences between the baseline SIF approach and the Concatenated approach. By the Precision-Recall metric, there were also significant differences between our proposed SiameseNet and the Concatenated approach.
Regarding the spaCy model, there were significant differences between the spaCy model performance compared to all methods that relied on our suggested time-based entity embedding or semi-supervised event approach.


\section{Discussion}

Upon analyzing the results presented in the previous section, we can conclude the following:
First, the utilization of entities, themes, and events to construct news embeddings has shown promising results, as reflected in Tables~\ref{tab:table_yearly_aggregated_stats} and~\ref{tab:summarized_reults}.
Even though the task of shared event detection between pairs of articles in a given timeframe is challenging, as discussed in the Introduction section, all proposed methods achieved extremely high ROC AUC scores, all above 0.95, while also providing impressive results concerning the more challenging metric of PR AUC.

Second, utilizing the semi-supervised approach, which considers cross-document latent information given the tagged events, contributes to our method performance. 
From Figure \ref{fig:box_plot_methods_comparison}, it is evident that the semi-supervised approach achieves significant improvement compared to the SIF baseline by the ROC AUC metric on all tasks. 
Additionally, from analyzing Figure \ref{fig:total_methods_comparison}, we see that even though there is a high correlation between the methods’ performance, a variance in performance exists. 
Namely, we found some cases where the baseline performed better than the semi-supervised approach. 
We attribute this behavior to two reasons: (a) we found a moderate correlation of 0.369 for each dataset size and the gap in performance between the SIF and semi-supervised approach; more data led to a bigger gap in favor of the Siamese network, (b) a lack of adaptation of the Siamese network to the given dataset or task led to variances in its results.

Third, our method, which generates news article embeddings, produced embeddings for articles and entities on a large scale of news data without GPU. Using the resources described in Appendix \ref{sec_appendix}, we ran every step of the pipeline and every experiment on all the generated datasets in parallel. 
For each entity and theme we monitored, over 320,000 overall (see Table \ref{tab:monitor_entity_type_counts}), we generated topical embeddings for every period they appeared in from 07/2015 to 01/2021.
We have considered all articles from dozens of monitored media sources and their affiliates, as presented in Figure \ref{fig:sources_volume}, resulting in a dataset of over 850,000 articles. This enables researchers with fewer resources to investigate large-scale online news using this method.

Fourth, we constructed topical monthly entity embeddings for over 320,000 distinct entities and themes. The pooling mechanism that controlled the training set of the GloVe algorithm served as a critical point in this research. 
This mechanism not only allowed us to control the resource utilization of our suggested method but also fine-tuned the representation of each entity to a recent time frame, such as its article’s publication date. 
In Table \ref{tab:entity_embed_example}, we further explain and demonstrate real-world examples of the importance of topical embeddings. We plan to share these topical entity embeddings with the research community. These topical embeddings could serve other studies examining how relationships between entities change over time.

Fifth, as presented in the previous section, the experimental results demonstrate the superiority of our proposed method compared to the baseline method SIF and the biggest CPU spaCy method. 
This can be observed in Table \ref{tab:summarized_reults}.
We detected only three instances out of 66 datasets and two tasks in which the SIF approach surpassed our technique. The spaCy model and the Semi-Supervised infused spaCy embeddings fell significantly short compared to all other approaches in all of the designed experiments and measured metrics. Additionally, Figure \ref{fig:box_plot_spacy_comparison} shows that our Semi-Supervised approach significantly improves the spaCy embeddings for news articles for the task of common event attribution.

Sixth, we observed that in the cases in which better performance is yielded by the Siamese network or SIF Methods compared to the Concatenated approach, there was a high gap between the SIF approach and the Siamese approach, leading to the failure of the concatenated approach to successfully account for additional information by utilizing both approaches and resulting in lesser performance than taking the best approach given the dataset.

Lastly, the method presented in this research has some limitations worth noting: First, it relies on the historical existence of event tagging for news articles, which might not always be available or reliable enough. Second, as suggested by the results, our methodology is relatively static, and the pipeline should preferably be manually tuned to provide adaptive embeddings to the required task; these concerns both in the semi-supervised training process and in the feature selection phase.


\section{Conclusions and Future Work}
\label{sec_conclusion}
This research has presented a novel approach for generating event-based embeddings for news articles, addressing the limitations of existing methods that rely solely on the textual content of documents, disregard runtime and memory costs, ignore the contextual time of article publication date, or do not rely on cross-document information. We do so by leveraging the historical context of entities, themes, and their associations with specific events. Our proposed method aimed to capture the latent relationships between these elements, ultimately enhancing the representation of news articles in a vector space.

We demonstrated our method’s ability to produce article embeddings that consider the article’s published time and represent the latent event information that appears in them, all while using relatively low resources on large-scale datasets. We also showed the method’s ability to surpass and improve existing solutions.

This research offers multiple directions for possible future work: The first direction is investigating the interpretability and explainability of the generated embeddings, which could be a valuable research direction. Understanding the factors that contribute to the embeddings’ representations and their relationships with events and entities could lead to more transparent and trustworthy models and facilitate the incorporation of human expertise and domain knowledge into the embedding generation process.

This also paves the way for analyzing the relationships between certain entities and tense or biased media coverage. In the future, we may apply a holistic approach to event clustering based on the methods proposed in this work to facilitate possible research in media slant or political news.

Another interesting opportunity for research could be to suggest an automated methodology for training a flexible semi-supervised network that might result in substantial improvement in the results, given the dataset it receives as input. This makes sense, considering the high variance of both dataset size and the positive rate as presented in depth in Table \ref{tab:datasets_statistics}.


\section*{Acknowledgments} 

\addcontentsline{toc}{section}{Acknowledgments} 

We would also like to thank Dr. Kalev Leetaru the Founder of the GDELT Project, for his assistance with understanding the GDLET project, and the possibilities its data contained.

In addition, while drafting this article, we used ChatGPT for editing according to necessity.
We also used Grammarly to check the grammar, spelling, and clarity of the text in this study.


\phantomsection
\bibliographystyle{unsrt}
\bibliography{thesis}

\clearpage
\appendix
\section{Research Notes}
\label{sec_appendix}
In the appendix, we present the detailed implementation notes of our method, including the running environment and the parameter settings for the deployed algorithms and Neural Networks.
Finally, we provide further references to the GDELT raw data.

\subsection{Implementation Notes}
\label{sec_deep_implementation_notes}
\begin{figure*}[b]
    \centering
    \vspace{0.5cm}
    \includegraphics[width=1\textwidth]{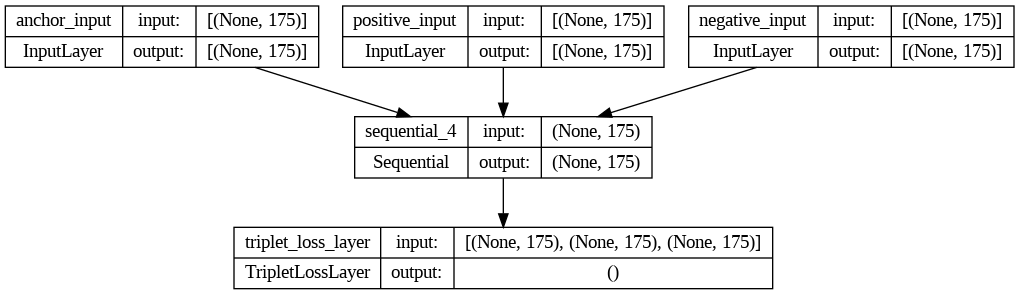}
    \caption[Siamese Network Parameters Archtecture]{This figure presents a summary of the proposed SiameseNet Architecture.}
    \label{fig:siamesenet_architecture}
\end{figure*}

\noindent \textbf{Running Environment.} 
The experiments are conducted on a single Linux server with AMD EPYC 7702P 64-Core Processor and
512Gb RAM, no GPU used for train or evaluation. 
Our methods are implemented on Tensorflow 2.4.0, Keras 2.7.0, and Python 3.8.5.
\\\\
\textbf{Hyperparameter settings.} Adam optimizer with a Learning Rate of 0.0015, The network architecture was a simple input and output layer of 175 units with three hidden layers of 128 units, we've also defined l2 regularization on top of each layer of the model. The models were trained on 10,000 epochs, and a batch size of 64, which consisted of 50 random pairings of articles and 14 hard pairings, as described by our defined triplets mining policy.
The architecture of the Siamese network is presented in Figure \ref{fig:siamesenet_architecture}.

\subsection{Data Notes}
\label{sec_additional_data_notes}
For further references on the GDELT raw data see the GDELT event database format codebook
\footnote{\url{http://data.gdeltproject.org/documentation/GDELT-Event\_Codebook-V2.0.pdf}}, and also GDELT Global Knowledge Graph (GKG)
data format codebook \footnote{\url{http://data.gdeltproject.org/documentation/GDELT-Global_Knowledge_Graph_Codebook-V2.1.pdf}}.

To provide a better indication of the pitfalls of news event association, we will provide real-world examples of articles that have been published, a short NER output on a sample of their content, and their event association by GDELT:
\begin{figure*}
\centering
\subfloat{\includegraphics[width=0.65\textwidth]{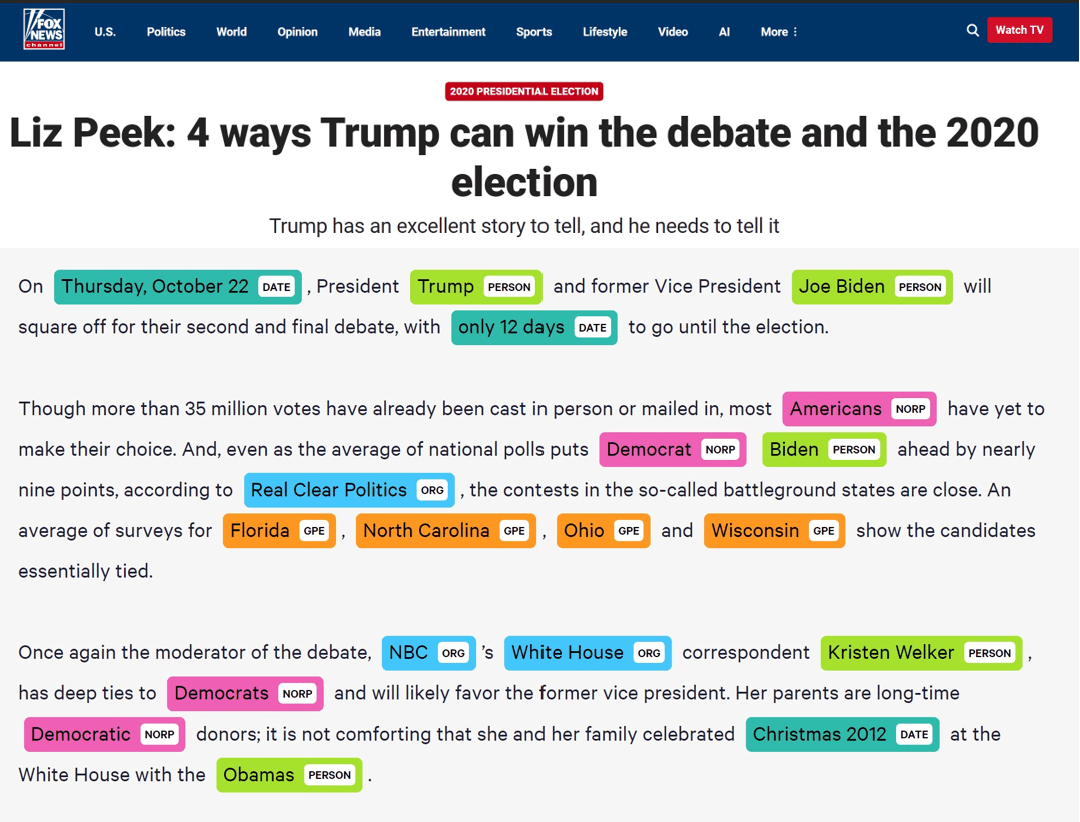}}\\
\subfloat{\includegraphics[width=0.65\textwidth]{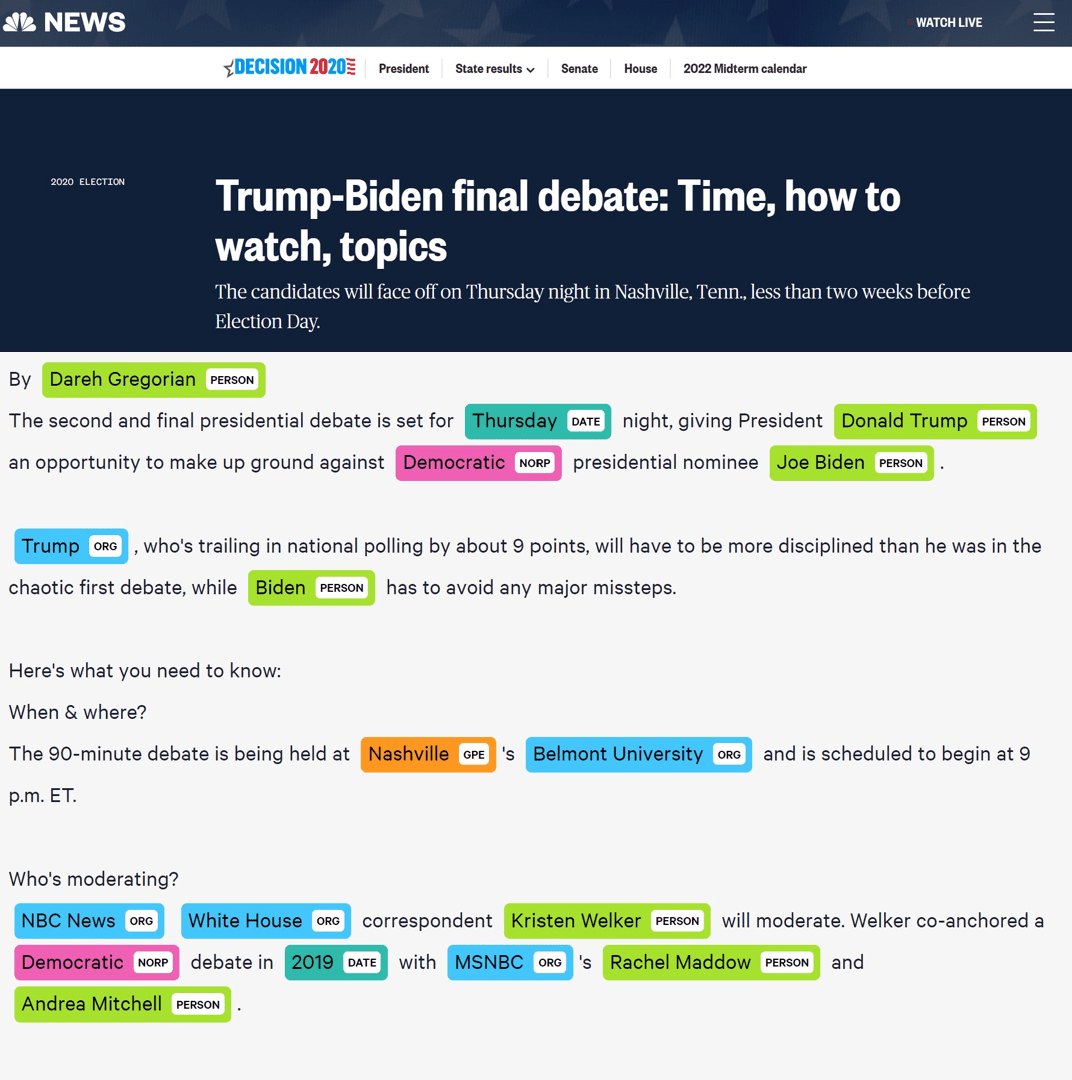}}
\caption[Real World Example of NER - US Debate Election]{These images provide examples of classic NER output on two US 2020 Election Debate News Articles}
\label{fig:2020_debate_articles}
\end{figure*}

\begin{figure*}
\centering
\subfloat{\includegraphics[width=0.65\textwidth]{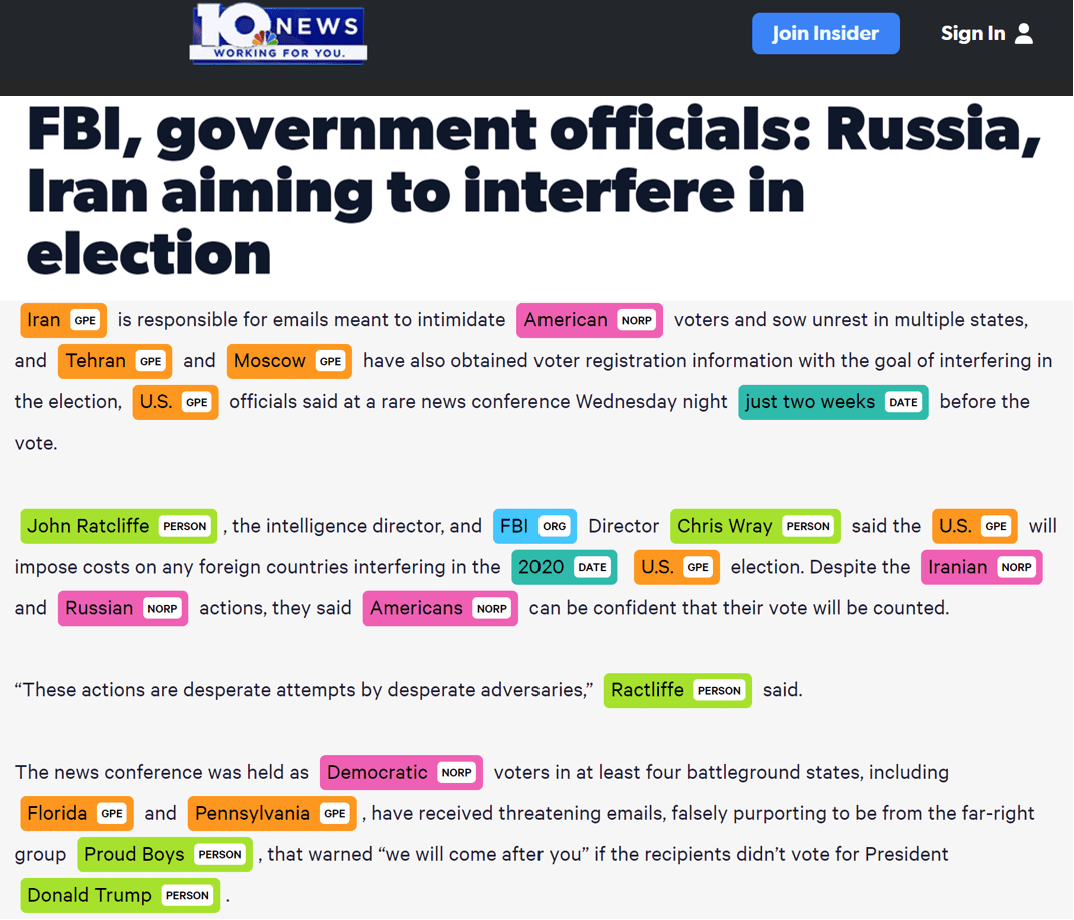}}\\
\subfloat{\includegraphics[width=0.65\textwidth]{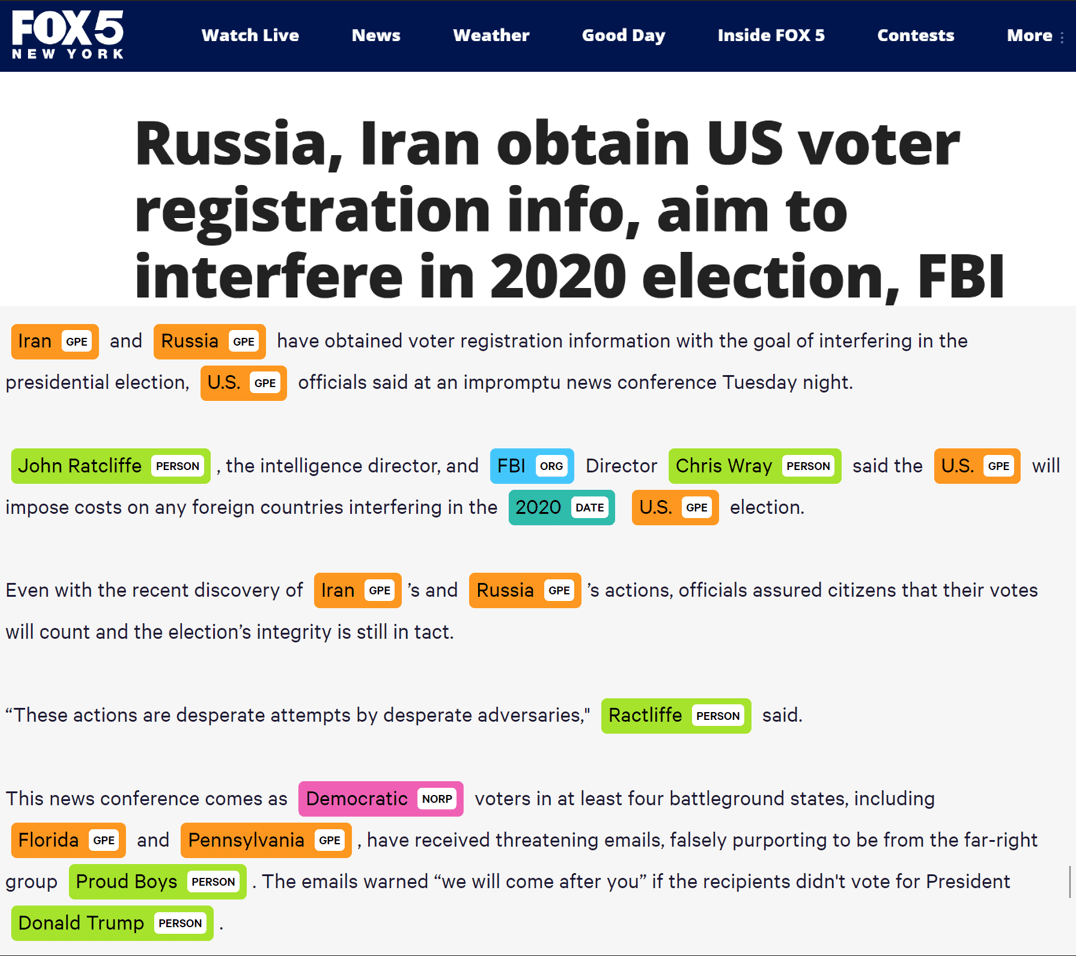}}
\caption[Real World Example of NER - Iran MEddling in US Election]{These images provide examples of classic NER output on two Iran Meddling with US 2020 Election News Articles}
\label{fig:Iran election intervention}
\end{figure*}

Examining the two articles in Figure \ref{fig:2020_debate_articles}, and the NER output on their content it seems fairly obvious that those discussed a common event: the upcoming second presidential debate. 
The logic of utilizing entities and themes to detect events is crystal clear here. 
As expected, the GDELT automated coding system agrees with our analysis, only it associated the NBC article with another event, that unfortunately is relevant only to this specific article in all of our collected datasets. 
On that note, one could claim that the Fox News article discussed another event: the White House 2012 Christmas Pary. 
This represents the problematic nature of event association and the difficulty behind it. 

Another real-world example is presented in Figure \ref{fig:Iran election intervention}, again the NER output of the articles' content marked in it.
Just like in the previous Figure \ref{fig:2020_debate_articles}, it seems that this pair of articles is extremely similar and therefore shares an event, we would label this event as something like 'Iran Meddling with 2020 US Election'. 
However, by GDELT annotations those articles share no common event, even though they share similar context, entities, and themes. 

{\subsubsection{Source Data Description}
\label{sec_dataset_description}
The following are the descriptions of the files generated by GDELT and used in this work, to construct the datasets.

\begin{itemize}
  \item \textbf{GKG Files}: These files contained information on each document, such as an identifier (URL), date of the article, source, and most importantly its entities, the entities we collected throughout this paper were - Themes, Persons, Organizations, Locations, and Dates.
  \item \textbf{Mentions Files}: These files connect between events and articles. They contain information on the relationships between them, such as the global event identifier, event time date, article time date, article identifier, the actors of the event, and so forth. They also contain the certainty score in the connection. We considered only mentions with absolute certainty by GDELT. 
  \item \textbf{Export Files}: These files contain information on the events. They contain information on the relationships between them, such as the global event identifier, event time date, article time date, article identifier, the actors of the event, and so forth.
\end{itemize}}

Table \ref{tab:datasets_statistics} provides a comprehensive overview of the attributes included in each monthly dataset that has been generated, from the GDELT source data, in the period of 01/2015 - 01/2022.

\subsection{Embedding Generation Notes}
\label{sec_Embedding_Generation_Notes}
In this section, we will provide examples of the effect our methodology had on the embedding generation process.

{
\footnotesize{
\begin{table*}[t]
\caption{This table, contains the closest neighbors of a given entity, by a trained GloVe model on month 01/2016, with and without the pooling mechanism.}
\vspace{0.5cm}
\begin{tabular}{c|c|c|c}
\multicolumn{1}{c|}{Entity}      & Entity Neighbor & GloVe                        & GloVe+Pool Mechanism         \\ \hline
\multirow{5}{*}{9.11}            & First            & ECON\_SOVEREIGN\_DEBT        & 11.9.2001                    \\
                                 & Second           & Marlon\_Brando               & World\_Trade\_Center         \\
                                 & Third            & Stockard\_Channing           & Memorial\_Museum             \\
                                 & Fourth           & Elizabeth\_Taylor            & WB\_2467\_TERRORISM          \\
                                 & Fifth             & Michael\_Jackson             & TAX\_FNCACT\_FIREMEN         \\ \hline
\multirow{5}{*}{Vladimir\_Putin} & First            & TAX\_ETHNICITY\_RUSSIANS     & TAX\_ETHNICITY\_RUSSIAN      \\
                                 & Second           & TAX\_ETHNICITY\_RUSSIAN      & TAX\_WORLDLANGUAGES\_RUSSIAN \\
                                 & Third            & TAX\_WORLDLANGUAGES\_RUSSIAN & RS                           \\
                                 & Fourth           & Alexander\_Litvinenko        & TAX\_ETHNICITY\_RUSSIANS     \\
                                 & Fifth             & RS                           & Alexander\_Litvinenko       
\end{tabular}
\label{tab:entity_embed_example}
\end{table*}}
}

We will investigate how the pooling mechanism affected the generation process of entity embeddings \ref{subsec_entity_embed}.
An example of the pooling effect is exemplified by the examples in Table \ref{tab:entity_embed_example}. 
The Table showcases two examples of how data reliant is the Glove model.
The single monthly trained GloVe model was not able to represent the 9.11 entity. 
It was, however able to successfully encapsulate the Vladimir\_Putin entity, while including Alexander\_Litvinenko as a close neighbor.

Alexander Litvinenko was a former officer of the Russian Federal Security Service (FSB) who defected and became a British citizen. He died suddenly from poisoning in 2006, leaving many questions unanswered. However, in January 2016, a public inquiry concluded that the two suspects who carried out Litvinenko's murder were likely acting under the direction of the FSB and with the approval of Russian President Vladimir Putin. The GloVe, both with and without the pooling mechanism, captured the topical relationship between Litvinenko and Putin during that period.

\subsection{Statistical Evaluation Notes}
\label{appendix_statistical_test_results}
To perform the statistical evaluation Friedman test was used, the test evaluated whether a significant difference in the result exists between each pair of methods.
The results are described in Table \ref{tab:friedman_results}

{\begin{table}[h]
\centering
\footnotesize 
\caption[Friedman Test Results per Task and Metric]{Friedman test results per context - task and metric}
\vspace{0.5cm}
\label{tab:friedman_results}
\begin{tabular}{P{1.5cm}|P{3cm}|P{1.75cm}}
Task    & Metric               & P-value  \\
\hline
Monthly & Precision-Recall AUC & 3.95E-14 \\
Monthly & ROC AUC              & 5.33E-21 \\
Daily   & Precision-Recall AUC & 3.94E-19 \\
Daily   & ROC AUC              & 3.18E-22
\end{tabular}
\end{table}
}

\makeatletter
\setlength{\@fptop}{0pt}
\makeatother

{
\onecolumn
{\begin{table*}[ht!]
\centering
\footnotesize 
\caption[Nemenyi Post-Hoc Test Results]{Nemenyi post-hoc test results per context - task and metric}
\vspace{0.5cm}
\begin{tabular}{P{1.5cm}|P{3cm}|P{2.5cm}|P{2.5cm}|P{2cm}}
Task& Metric  & Method 1  & Method 2  & P-value \\
\hline
\multirow{3}{*}{Monthly} & \multirow{3}{*}{Precision-Recall AUC} 
 & Concatenation & SiameseNet & 0.001\\
 && Concatenation & SIF& 0.001\\
 && SiameseNet & SIF& 0.8447\\
\hline
\multirow{3}{*}{Monthly} & \multirow{3}{*}{ROC AUC}  & Concatenation & SiameseNet & 0.8447  \\
 && Concatenation & SIF& 0.001\\
 && SiameseNet & SIF& 0.001\\
 \hline
\multirow{3}{*}{Daily}& \multirow{3}{*}{Precision-Recall AUC} & Concatenation & SiameseNet & 0.001\\
 && Concatenation & SIF& 0.001\\
 && SiameseNet & SIF& 0.8945  \\
 \hline
\multirow{3}{*}{Daily}& \multirow{3}{*}{ROC AUC}  & Concatenation & SiameseNet & 0.9  \\
 && Concatenation & SIF& 0.001\\
 && SiameseNet & SIF& 0.001  
\end{tabular}
\label{tab:Nemenyi_results}
\end{table*}}

\footnotesize
\begin{longtable}
{P{1cm}P{1cm}|P{2cm}P{2cm}P{2cm}P{2.5cm}P{2.5cm}}
\caption[Full Datasets Statistics Description]{Datasets statistics are used in this work for performance analysis.}
\label{tab:datasets_statistics}\\
\toprule
\multicolumn{2}{c}{\textbf{Dataset}} & \multicolumn{5}{c}{\textbf{Statistics}} \\
\cmidrule(rl){1-2} \cmidrule(rl){3-7}
Year& Month& \#Articles& \#Events& \#Mentions& \%Monthly PR& \%Daily PR\\
\hline
 \multirow{6}{2em}{2015}
& 7 & 14961& 18746& 69644& 0.074\% & 0.637\%\\
& 8 & 12872& 16743& 58580& 0.052\% & 0.496\%\\
& 9 & 13701& 16772& 64938& 0.065\% & 0.528\%\\
& 10& 11723& 14013& 51964& 0.040\% & 0.361\%\\
& 11& 13389& 15968& 61664& 0.057\% & 0.512\%\\
& 12& 12231& 14541& 54358& 0.040\% & 0.473\%\\

 \hline
 \multirow{12}{2em}{2016}
& 1 & 12880& 14809& 55648& 0.040\% & 0.340\%\\
& 2 & 13763& 16514& 60482& 0.042\% & 0.353\%\\
& 3 & 14316& 17174& 60814& 0.037\% & 0.327\%\\
& 4 & 13285& 15285& 53896& 0.037\% & 0.301\%\\
& 5 & 14139& 16923& 57166& 0.050\% & 0.402\%\\
& 6 & 14860& 18055& 62852& 0.051\% & 0.414\%\\
& 7 & 15484& 17932& 71014& 0.035\% & 0.355\%\\
& 8 & 12916& 15085& 52476& 0.042\% & 0.320\%\\
& 9 & 15010& 16837& 61678& 0.045\% & 0.369\%\\
& 10& 13566& 16402& 55620& 0.036\% & 0.324\%\\
& 11& 13498& 16492& 57346& 0.054\% & 0.454\%\\
& 12& 13037& 15208& 53336& 0.039\% & 0.397\%\\

 \hline
 \multirow{12}{2em}{2017}
& 1 & 13989& 17375& 59614& 0.041\% & 0.328\%\\
& 2 & 12799& 15761& 51010& 0.039\% & 0.304\%\\
& 3 & 15070& 18638& 63242& 0.042\% & 0.375\%\\
& 4 & 14166& 17521& 61750& 0.042\% & 0.331\%\\
& 5 & 14880& 18419& 64636& 0.040\% & 0.358\%\\
& 6 & 13616& 16399& 58848& 0.039\% & 0.364\%\\
& 7 & 12376& 14882& 53130& 0.055\% & 0.407\%\\
& 8 & 13325& 16253& 55712& 0.062\% & 0.524\%\\
& 9 & 13458& 16543& 59196& 0.040\% & 0.366\%\\
& 10& 13996& 16977& 61068& 0.034\% & 0.305\%\\
& 11& 13108& 16295& 57012& 0.041\% & 0.357\%\\
& 12& 11481& 14407& 49094& 0.048\% & 0.498\%\\

 \hline
 \multirow{12}{2em}{2018}
& 1 & 12445& 15108& 51280& 0.045\% & 0.370\%\\
& 2 & 11923& 14758& 49846& 0.049\% & 0.355\%\\
& 3 & 15170& 18369& 63828& 0.040\% & 0.365\%\\
& 4 & 14326& 17520& 63204& 0.047\% & 0.416\%\\
& 5 & 15436& 19194& 68934& 0.048\% & 0.443\%\\
& 6 & 14167& 17737& 61436& 0.051\% & 0.410\%\\
& 7 & 13790& 16397& 59174& 0.049\% & 0.408\%\\
& 8 & 15655& 17846& 68822& 0.045\% & 0.395\%\\
& 9 & 12070& 13728& 48196& 0.038\% & 0.318\%\\
& 10& 13047& 15031& 53366& 0.055\% & 0.452\%\\
& 11& 11933& 14268& 49990& 0.039\% & 0.349\%\\
& 12& 10543& 12371& 43796& 0.058\% & 0.598\%\\

 \hline
 \multirow{12}{2em}{2019}
& 1 & 11192& 13056& 45370& 0.040\% & 0.335\%\\
& 2 & 10940& 12563& 42594& 0.038\% & 0.291\%\\
& 3 & 13204& 14906& 54660& 0.036\% & 0.327\%\\
& 4 & 12754& 14808& 52608& 0.043\% & 0.329\%\\
& 5 & 13510& 15486& 53748& 0.038\% & 0.348\%\\
& 6 & 12755& 13972& 51938& 0.042\% & 0.375\%\\
& 7 & 10657& 11824& 40732& 0.049\% & 0.407\%\\
& 8 & 9938 & 11191& 39042& 0.047\% & 0.437\%\\
& 9 & 10332& 11711& 41114& 0.038\% & 0.315\%\\
& 10& 11544& 12573& 45260& 0.031\% & 0.275\%\\
& 11& 12163& 13089& 46430& 0.036\% & 0.328\%\\
& 12& 10834& 12218& 41796& 0.040\% & 0.451\%\\

 \hline
 \multirow{12}{2em}{2020}
& 1 & 12432& 14406& 51486& 0.063\% & 0.480\%\\
& 2 & 12404& 14033& 50100& 0.033\% & 0.280\%\\
& 3 & 14105& 16033& 58402& 0.038\% & 0.369\%\\
& 4 & 11814& 13644& 46596& 0.030\% & 0.271\%\\
& 5 & 11658& 13370& 45256& 0.047\% & 0.377\%\\
& 6 & 13235& 15100& 52990& 0.034\% & 0.292\%\\
& 7 & 14963& 17008& 57558& 0.026\% & 0.233\%\\
& 8 & 13764& 14974& 53332& 0.031\% & 0.277\%\\
& 9 & 12427& 13660& 46116& 0.025\% & 0.248\%\\
& 10& 6769 & 7649 & 25374& 0.044\% & 0.441\%\\
& 11& 6177 & 6825 & 22626& 0.052\% & 0.307\%\\
& 12& 12521& 13572& 47118& 0.034\% & 0.328\%\\

\hline
\end{longtable}
\twocolumn}

\end{document}